%% file: arxiv.tex
\documentclass[letterpaper]{article}
\usepackage[preprint]{aaai2027}

\usepackage[hyphens]{url}
\usepackage{graphicx}
\usepackage{natbib}
\usepackage{caption}
\usepackage{algorithm}
\usepackage{algorithmic}

\usepackage{amsmath}
\usepackage{amssymb}
\usepackage{amsthm}
\usepackage{xcolor}
\definecolor{PaDocDeepGreen}{RGB}{0,100,0}
\usepackage{placeins}

\usepackage{newfloat}
\usepackage{listings}
\DeclareCaptionStyle{ruled}{labelfont=normalfont,labelsep=colon,strut=off}
\floatstyle{ruled}
\newfloat{listing}{tb}{lst}{}
\floatname{listing}{Listing}

\usepackage{booktabs}
\usepackage{multirow}
\usepackage{makecell}

\newcommand{\isArxivVersion}{}

\input{title}
\author{
  Hao Yu\textsuperscript{\rm 1,\rm 2}\equalcontrib,
  Jiabo Zhan\textsuperscript{\rm 1}\equalcontrib,
  Kang Liu\textsuperscript{\rm 2},
  Linnan Zhao\textsuperscript{\rm 2},
  Dongxu Yue\textsuperscript{\rm 1},
  Rui Chen\textsuperscript{\rm 1},\\
  Jinglin Wang\textsuperscript{\rm 1},
  Chong Sun\textsuperscript{\rm 2},
  Chen Li\textsuperscript{\rm 2},
  Jing LYU\textsuperscript{\rm 2},
  Chun Yuan\textsuperscript{\rm 1}\corresponding
}
\affiliations{
  \textsuperscript{\rm 1}Tsinghua University\quad
  \textsuperscript{\rm 2}Wechat Vision, Tencent
}

\begin{document}

\input{preamble}

\maketitle

\input{sec/0_abstract}

\input{sec/1_introduction}

\input{sec/2_related_work}

\input{sec/3_method}

\input{sec/4_experiments}
\input{sec/5_conclusion}

\bibliography{main}

\clearpage
\appendix
\setcounter{secnumdepth}{2}
\input{sec/A_derivations}
\input{sec/B_pageiou}
\input{sec/C_ablations_of_training_paradigm}
\input{sec/D_additional_analysis}

\input{sec/E_additional_serving_metrics}

\end{document}

%% file: title.tex
\title{PaDoc: Layout-Grounded Parallel Decoding for Document Parsing}

%% file: preamble.tex
\newcommand{\red}[1]{{\color{red}#1}}
\newcommand{\orange}[1]{{\color{orange}#1}}
\newcommand{\green}[1]{{\color{green}#1}}

\newcommand{\todo}[1]{{\color{red}[todo] #1}}
\newcommand{\wip}[1]{{\color{orange}[wip] #1}}
\newcommand{\update}[1]{{\color{cyan}[update] #1}}
\newcommand{\done}[1]{{\color{green}[done] #1}}


\theoremstyle{definition}
\newtheorem{assumption}{Assumption}

%% file: sec/0_abstract.tex
\begin{abstract}
End-to-end document parsers provide a unified interface, but
serialize page layouts and regional contents into one autoregressive sequence.
This formulation forces independent regions onto a decoding path whose
length grows with the total content, whereas crop-based two-stage parsers expose
region-level parallelism at the cost of repeated visual prefills and fragmented
page context. To retain full-page context while removing
dependencies, we propose \textsc{PaDoc}, a layout-grounded parser that treats
the predicted layout as a branching structure over a shared page
representation. Under a region-sufficiency assumption, we derive a
prefix-conditioned factorization in which the layout stream and regional
content branches advance concurrently, reducing the decoding depth to
the longest layout-content path. We realize this factorization within a single
MLLM: packed variable-length ancestor attention preserves the
visibility under standard next-token training, while masked parallel decoding
creates branches that the evaluated vLLM backend serves as concurrent
requests with cache-resident shared-prefix reuse. On OmniDocBench Full,
\textsc{PaDoc} attains an Overall layout F1 of 91.1 and, among end-to-end
parsers, a top-tier Overall score of 94.24 together with the best Text Edit
(0.038) and Formula CDM (95.59). On a 384-page subset and one A800 GPU, it
is the fastest end-to-end parser at five concurrency levels,
improving valid-page throughput by 67.4--118\% and reducing P95 latency by
39.2--54.9\% relative to a same-backbone Sequential SFT baseline.
\ifdefined\isArxivVersion
Code is available at \url{https://github.com/Longin-Yu/Padoc}.
\fi
\end{abstract}

%% file: sec/1_introduction.tex
\section{Introduction}

Document parsing transforms page images into structured representations of
layout regions, reading order, text, formulas, tables, and figures. Multimodal
systems increasingly cast this task as sequence generation, using a unified
autoregressive model to emit Markdown, HTML, JSON, or layout-content records.
Early image-to-sequence models established this paradigm
\citep{kim2022donut,lee2023pix2struct,blecher2024nougat}, and subsequent document
VLMs extended it to full-page structured parsing
\citep{nassar2025smoldocling,poznanski2025olmocr,niu2025mineru2,dong2026qianfan}.
Although this unified interface reduces task-specific modeling, it retains the
sequential execution semantics of autoregressive decoding.

\begin{figure*}[!t]
\centering
\includegraphics[width=\textwidth]{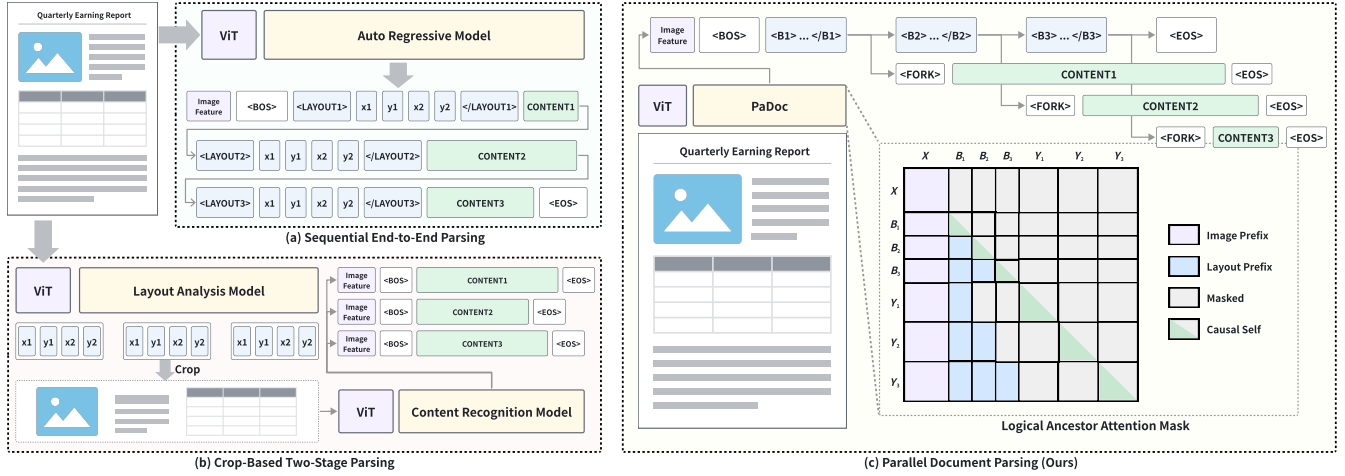}
\caption{Comparison of three document-parsing paradigms. (a) Sequential
end-to-end parsing serializes all layout and content tokens after a single
visual encoding; (b) crop-based two-stage parsing exposes region-level
parallelism but requires crop extraction and repeated visual prefills.
(c) \textsc{PaDoc} forks content branches from shared full-page image and layout
prefixes. The displayed ancestor mask is logical: training realizes it with
packed variable-length attention rather than materializing a dense mask, while
serving instantiates concurrent streams with shared-prefix cache reuse.}
\label{fig:main-overview}
\end{figure*}

As illustrated in Figure~\ref{fig:main-overview}(a), a flat output sequence does
not reflect the conditional structure of a document. Consider a serialization
$(b_1,y_1,b_2,y_2,\ldots,b_N,y_N)$, where $b_k$ specifies the layout of region
$k$ and $y_k$ denotes its category, text, formula, or table representation. A
causal decoder must complete $y_k$ before generating $b_{k+1}$, irrespective of
whether the two regions are spatially separated and semantically independent.
The resulting critical path therefore scales with the sum of all regional
content lengths and becomes particularly costly for pages containing many
independently recognizable regions.
Layout-first systems expose region-level parallelism, as shown in
Figure~\ref{fig:main-overview}(b), but their per-region crops require repeated
visual encoding and restrict access to page-level context
\citep{cui2026paddleocr,niu2025mineru2,feng2025dolphin,feng2026dolphin}.
General accelerators instead create speculative or textual branches that are
not necessarily grounded in visual document instances and may be excluded from
the final output
\citep{leviathan2023fast,cai2024medusa,li2024eagle,liao2026hsd,yao2025deft}.

These limitations, which respectively fragment visual context and introduce
branches ungrounded in document layout, motivate a formulation that derives
parallelism from the predicted layout while preserving a shared page
representation. To this end, we propose \textsc{PaDoc}, a layout-grounded
document parser implemented within a single MLLM, as illustrated in
Figure~\ref{fig:main-overview}(c). \textsc{PaDoc} adopts a region-specific
conditional-independence assumption under which the content $Y_k$ of region
$k$ is independent of other regions' contents conditioned on the page image
$X$ and its layout $B_k$. This assumption yields a prefix-conditioned
factorization in which the layout stream predicts $B_{k+1}$ and the
corresponding content branch predicts $Y_k$ from the common prefix
$(X,B_{\le k})$. The two factors can consequently be decoded concurrently.
Unlike crop-based recognition, every content branch retains the complete page
image as a shared visual prefix and differs only in the visible layout prefix.

\textsc{PaDoc} realizes this factorization without separate detector or
recognizer heads. During training, logical ancestor attention exposes the image
and appropriate layout ancestors to each branch while excluding sibling
contents. Its packed variable-length implementation avoids a dense mask while
retaining the standard next-token objective. At inference, masked parallel
decoding uses region-closing tokens to instantiate branch-isolated content
streams under the same ancestor visibility. Our evaluated vLLM
backend~\citep{kwon2023efficient} realizes these streams as concurrent requests
with cache-resident image and layout prefix reuse. The resulting logical
decoding depth is the longest root-to-leaf layout-content path rather than the
sum of all regional content lengths. We initialize the model
directly from the original Qwen3-VL-2B checkpoint~\citep{bai2025qwen30vl} and adapt
all modules through continual pre-training (CPT) followed by high-quality
supervised fine-tuning (SFT).

We evaluate layout analysis, end-to-end parsing, and inference efficiency on
OmniDocBench~\citep{ouyang2024omnidocbench0}. \textsc{PaDoc} attains an Overall layout
F1 of 91.1. Among end-to-end parsers, it achieves a top-tier Overall
score of 94.24, together with the best Text Edit and Formula CDM. On an A800 GPU, it's the fastest end-to-end parser at all five tested
concurrency levels. Relative to the same-backbone Sequential SFT baseline,
\textsc{PaDoc} improves throughput by
$67.4\sim118$\% and reduces P95 latency by $39.2\sim54.9$\%. Despite its 2.1B size, it
also surpasses the 1.0B HunyuanOCR-1.5 at every concurrency and substantially
outpaces the 0.7B MonkeyOCRv2, establishing a favorable quality-efficiency
trade-off.
Our contributions are threefold:
\begin{itemize}
    \item We derive a prefix-conditioned factorization of document parsing that
    exposes region-level parallelism while preserving a shared page-image
    prefix, thereby reducing the logical decoding depth to the longest
    layout-content path.
    \item We develop a single-model training formulation based on ancestor
    attention and an efficient packed tree-varlen realization, retaining
    standard next-token training without auxiliary prediction heads or losses.
    \item We implement layout-grounded parallel serving with shared-prefix KV
    reuse and demonstrate a favorable quality--efficiency trade-off against
    representative end-to-end and two-stage document parsers.
\end{itemize}

%% file: sec/2_related_work.tex
\section{Related Work}


\paragraph{Document parsing as multimodal generation.}
Recent document parsing systems increasingly formulate document understanding as multimodal generation, directly converting document images or PDF pages into Markdown, structured text, or task-specific markup. 
Compared with OCR-centric pipelines, this paradigm offers a unified output space for heterogeneous elements such as text, tables, formulas, figures, and layout structures. 
Representative image-to-sequence methods, including Donut~\citep{kim2022donut}, Pix2Struct~\citep{lee2023pix2struct}, and Nougat~\citep{blecher2024nougat}, have demonstrated the effectiveness of generative document modeling, while recent document VLMs such as SmolDocling~\citep{nassar2025smoldocling} and olmOCR~\citep{poznanski2025olmocr} extend this paradigm to page-level structured parsing. 
Qianfan-OCR~\citep{dong2026qianfan} further unifies layout analysis, parsing, and understanding, with Layout-as-Thought highlighting the need for explicit layout grounding in end-to-end generation.
However, most generation-based parsers still serialize the entire page into a single autoregressive sequence, which imposes artificial sequential dependencies among spatially separated regions and becomes inefficient for dense, table-heavy, or text-rich documents.

\paragraph{Layout detection and instance-level recognition.}
Another line of work adopts a layout-first pipeline, where page elements are first localized and then recognized at the region level.
Recent detectors such as DocLayout-YOLO~\citep{zhao2024doclayout} and PP-DocLayout~\citep{sun2025pp} provide efficient structural front-ends, while systems such as DocBed~\citep{zhu2022docbed}, Dolphin~\citep{feng2025dolphin, feng2026dolphin}, and HSD~\citep{liao2026hsd} further exploit layout-guided decomposition for local or parallel document parsing.
Related instance-level recognition methods such as SPTS v2~\citep{liu2023spts} also suggest the efficiency benefits of separating localization from recognition.
However, when region recognition is performed largely independently, such pipelines may weaken page-level context, including reading order, section hierarchy, table-caption relations, and cross-region references. This creates a tension between local parallelism and global document coherence.

\paragraph{Structured parallel generation.}
Structured generation represents outputs as sets, trees, graphs, or schemas rather than flat token sequences, enabling decoding to operate on meaningful units instead of individual tokens.
Recent methods exploit such structure for parallel decoding: SoT~\citep{ning2024skeleton} expands different skeleton points in parallel, APAR~\citep{liu2024apar0} learns control tokens that fork hierarchical autoregressive branches, and methods such as PASTA~\citep{jin2025learning}, ASPD~\citep{chen2025aspd}, and SPRINT~\citep{biju2026sprint} introduce learned chunk annotations, adaptive serial-parallel switching, or planning-based parallel execution.
System-level works such as Hogwild! Inference~\citep{rodionov2026hogwild} and DeFT~\citep{yao2025deft} further provide runtime support for shared attention caches and tree-structured inference.
While these methods show that explicit output structure can unlock parallel decoding, their structures are usually defined over text, reasoning steps, or abstract trees. Dense document parsing requires such parallel structures to be grounded in visual layout instances while still sharing page-level context.

%% file: sec/3_method.tex
\section{Method}

\newcommand{\model}{\textsc{PaDoc}}
\newcommand{\Crop}{\operatorname{Crop}}
\newcommand{\pa}{\operatorname{pa}}
\newcommand{\Anc}{\operatorname{Anc}}
\newcommand{\EOS}{\mathrm{EOS}}

\input{sec/3_1_formulation}

\input{sec/3_2_train}

\input{sec/3_3_infer}

%% file: sec/3_1_formulation.tex
\subsection{Problem Formulation}

\paragraph{Task formulation of document parsing.}
Let $X$ denote a document image, and let $x$ be an observed realization of $X$.
Document parsing aims to predict a collection of layout-content pairs
$
    \{(B_k,Y_k)\}_{k=1}^{N}
$,
where $B_k$ denotes the $k$-th layout region, e.g., a bounding box, and $Y_k$
denotes the textual or structured content associated with that region. We use
lowercase letters $b_k$ and $y_k$ to denote concrete realizations of $B_k$ and
$Y_k$, respectively. The regions are arranged in a canonical
order, such as reading order.
The target distribution of document parsing is
\begin{equation}
    p^\star(b_{1:N},y_{1:N}\mid x),
\end{equation}
By the chain
rule, it can be decomposed into a layout component and a content component:
\begin{equation}
    p^\star(b_{1:N},y_{1:N}\mid x)
    =
    p^\star(b_{1:N}\mid x)\,
    p^\star(y_{1:N}\mid x,b_{1:N}).
\end{equation}
The first term corresponds to layout analysis, while the second term corresponds
to content recognition conditioned on the layout.

\noindent
Both terms can be further factorized by the chain rule as
\begin{equation}
\delimiterfactor=650
\delimitershortfall=2em
\left\{
\begin{array}{@{}l@{}}
\displaystyle\;p^\star(b_{1:N}\mid x) = \prod_{k=1}^{N} p^\star(b_k\mid x,b_{<k}), \\[-.9em]
\displaystyle\;p^\star(y_{1:N}\mid x,b_{1:N}) = \prod_{k=1}^{N} p^\star(y_k\mid x,b_{1:N},y_{<k}).
\end{array}
\right.
\end{equation}
Given the canonical region order, the layout factor is a standard
autoregressive layout predictor. The main design choice is the content factor:
which parts of the full layout and content history are actually needed to
predict each $Y_k$. Thus, the central modeling problem is how to factorize and
parameterize
\begin{equation}
    p^\star(y_k\mid x,b_{1:N},y_{<k}).
\end{equation}
\paragraph{Native structural properties of document parsing.}
Each content variable $Y_k$ is grounded in a specific visual region $B_k$. We
capture this region-anchored structure with two conditional-independence
assumptions, stated using conditional mutual information~\citep{dobrushin1963general,wyner1978definition}.
Here $I(A;B\mid C)$ denotes the conditional mutual information between $A$ and
$B$ given $C$.

\begin{assumption}[Region-specific content sufficiency]
\label{asm:region-specific}
Given the document image $X$ and the current region $B_k$, other layout regions
and previously decoded contents provide no additional information about $Y_k$:
\begin{equation}
    I\!\left(
        Y_k;\,B_{<k},B_{>k},Y_{<k}
        \mid X,B_k
    \right)
    = 0.
\end{equation}
\end{assumption}

\begin{assumption}[Crop sufficiency]
\label{asm:crop}
For a region $B_k$, let $X_k=\Crop(X,B_k)$ denote its cropped image. The crop
preserves all visual evidence needed to recognize the content associated with
that region:
\begin{equation}
    I(Y_k;X,B_k\mid X_k) = 0.
\end{equation}
\end{assumption}

\paragraph{Crop-based two-stage formulation.}
A crop-based two-stage parser~\citep{cui2026paddleocr,niu2025mineru2} first
predicts the layout and then recognizes each region independently from
$x_k=\Crop(x,b_k)$. Applying Assumptions~\ref{asm:region-specific}
and~\ref{asm:crop} yields the following parameterization, which exposes the
same region-level parallelism:
\begin{equation}
    p_{\theta}^{\mathrm{crop}}(b_{1:N},y_{1:N}\mid x)
    =
    p_{\theta}^{\mathrm{B}}(b_{1:N}\mid x)
    \prod_{k=1}^{N}
    p_{\theta}^{\mathrm{Y}}(y_k\mid x_k).
\end{equation}
However, this formulation requires per-region crop extraction and repeated
visual prefills. Crop materialization introduces preprocessing and data-movement
overhead, while encoding each crop separately prevents the reuse of a shared
page-level visual prefix.

\paragraph{Shared-prefix parameterization of \model{}.}
Our goal is to retain the same conditional-independence structure, and thus the
region-level parallelism it enables, while removing these condition-switching
costs. The key is to replace the region-specific condition $x_k$ with a shared
condition that is a prefix of one common autoregressive context. \model{} keeps
the document image as a shared visual prefix and conditions the $k$-th content
branch on the generated layout prefix $B_{\le k}$, so the branch starts from
$(X,B_{\le k})$ rather than from a separately materialized crop $X_k$.
Consequently, all content branches share the visual prefill and differ only in
how much layout prefix they expose. It remains to justify that this
shared-prefix condition is sufficient for the same parallel decoding structure.
By the chain rule of conditional mutual information,
Assumption~\ref{asm:region-specific} implies the prefix-conditioned form (see
Appendix~\ref{sec:proof-prefix-cond} for a derivation):
\begin{equation}
    I\!\left(
        Y_k;\,B_{>k},Y_{<k}
        \mid X,B_{\le k}
    \right)
    = 0.
\end{equation}
This removes the dependence on future layout regions and previously decoded
contents once $(X,B_{\le k})$ is known:
\begin{equation}
    p^\star(y_k\mid x,b_{1:N},y_{<k})
    =
    p^\star(y_k\mid x,b_{\le k}).
\end{equation}
Substituting this identity into the exact decomposition yields the
prefix-conditioned factorization
\begin{equation}
\resizebox{0.88\linewidth}{!}{$\displaystyle
p^\star(b_{1:N},y_{1:N}\mid x)
=
\prod\limits_{k=1}^{N}
p^\star(b_k\mid x,b_{<k})\,
\prod\limits_{k=1}^{N}
p^\star(y_k\mid x,b_{\le k}).
$}
\end{equation}
Accordingly, \model{} parameterizes the document parsing distribution with a
layout stream and prefix-conditioned content branches:
\begin{equation}
\resizebox{0.84\linewidth}{!}{$\displaystyle
\begin{aligned}
    &p_{\theta}^{\model}(b_{1:N},y_{1:N}\mid x)             \\[-.4em]
    &=
    \prod_{k=1}^{N}
    p_{\theta}^{\mathrm{B}}(b_k\mid x,b_{<k})\,
    \prod_{k=1}^{N}
    p_{\theta}^{\mathrm{Y}}(y_k\mid x,b_{\le k})            \\[-.8em]
    &=
    p_{\theta}^{\mathrm{B}}(b_1\mid x)
    \prod_{k=1}^{N}
    \underbrace{
        p_{\theta}^{\mathrm{B}}(b_{k+1}\mid x,b_{\le k})\,
        p_{\theta}^{\mathrm{Y}}(y_k\mid x,b_{\le k})
    }_{\text{
        \small 
        share prefix $(x,b_{\le k})$}},
\end{aligned}
$}
\end{equation}
where we reindex the layout product and pair each next-region predictor with the
current-region content predictor, so that both factors at step $k$ are
conditioned on the same prefix $(x,b_{\le k})$. The boundary term
$b_{N+1}=\EOS$ denotes the end-of-sequence layout token that terminates the
layout stream.
This factorization enables parallel decoding: after the layout prefix
$B_{\le k}$ has been generated, the next layout region and the current region
content are both conditionally available,
\begin{equation}
\label{eq:parallel-decode}
    B_{k+1}
    \sim
    p_{\theta}^{\mathrm{B}}(\cdot\mid X,B_{\le k}),
    \qquad
    Y_k
    \sim
    p_{\theta}^{\mathrm{Y}}(\cdot\mid X,B_{\le k}).
\end{equation}
Since neither variable conditions on the other in the model factorization, they
can be decoded concurrently.

\noindent
Let $\ell(\cdot)$ denote the number of autoregressive decoding steps required to
generate its argument. A fully sequential layout-then-content decoder has
sequential depth
\begin{equation}
    D_{\mathrm{seq}}
    =
    \ell(B_{1:N})
    +
    \sum_{k=1}^{N}
    \ell(Y_k).
\end{equation}
In contrast, under ideal parallel scheduling, \model{} has critical-path decoding
depth
\begin{equation}
\resizebox{0.88\columnwidth}{!}{$\displaystyle
\begin{aligned}
    D_{\model}
    &=
    \max\left\{
        \ell(B_{1:N}),
        \max_{1\le k\le N}
        \bigl[
            \ell(B_{\le k})+\ell(Y_k)
        \bigr]
    \right\} \\
    &= \max_{1\le k\le N}
        \left\{
            \ell(B_{\le k})+\ell(Y_k)
        \right\}
\end{aligned}
$}
\end{equation}
Therefore, \model{} reduces the sequential dependency path from the total length
of all layout and content sequences to the longest path in the layout-content
dependency graph.

%% file: sec/3_2_train.tex
\subsection{Training Paradigm}
\label{sec:training-paradigm}

The prefix-conditioned factorization derived above
can be learned in a single forward pass over one shared sequence, without ever
materializing per-region crops, provided the attention pattern exposes the right
dependencies. \model{} realizes this with a purpose-built ancestor-attention
mask, so that a standard next-token objective on the packed sequence reproduces
the parallel graph structure at training time.
Inspired by APAR-style ancestor attention~\citep{liu2024apar0}, \model{} represents layout and
content variables as nodes in a directed acyclic graph. The graph contains the
image node $X$, layout nodes $B_1,\ldots,B_N$, and content nodes
$Y_1,\ldots,Y_N$, with ancestor sets
\begin{equation}
    \Anc(B_k)=\{X,B_{<k}\},
    \qquad
    \Anc(Y_k)=\{X,B_{\le k}\}.
\end{equation}
The \model{} graph in Figure~\ref{fig:main-overview}(c) follows these ancestor
sets. During training, each node attends only to its ancestors and to earlier
positions within the same node under a causal mask.
For a node $V$, tokens can attend to all tokens in $\Anc(V)$ and to non-future
tokens within $V$. Because $\Anc(B_k)=\{X,B_{<k}\}$ and
$\Anc(Y_k)=\{X,B_{\le k}\}$, this pattern exactly realizes the prefix-conditioned
factors $p_{\theta}^{\mathrm{B}}(b_k\mid x,b_{<k})$ and
$p_{\theta}^{\mathrm{Y}}(y_k\mid x,b_{\le k})$,
while preventing leakage from content
nodes to future layout nodes or to other content nodes.

\paragraph{Efficient ancestor attention.}
A direct realization materializes the ancestor mask as a dense
$S\times S$ tensor and therefore cannot use standard causal
FlashAttention. We instead express the same visibility pattern as a collection
of variable-length causal attention problems. The layout stream forms one
causal sequence, while the query tokens of content branch $Y_k$ attend to the
virtual key/value sequence $[X,B_{\le k},Y_k]$. Only the already-projected
key/value rows of the shared prefix are gathered; queries and token projections
are computed once, and gradients to reused prefix rows are accumulated by the
gather operation. This tree-varlen realization avoids quadratic mask storage
and retains standard causal FlashAttention kernels. We use it for all
experiments. Appendix~\ref{sec:training-paradigm-ablation} compares dense SDPA
and Flex Attention baselines with tree-varlen.

\noindent
Given ground-truth annotations
$(b_{1:N}^{\star},y_{1:N}^{\star})$, \model{} is trained with teacher forcing by
minimizing the negative log-likelihood
\begin{equation}
\resizebox{0.88\columnwidth}{!}{$\displaystyle
    \mathcal{L}(\theta)
    =
    -\sum_{k=1}^{N}\log
    p_{\theta}^{\mathrm{B}}\!\left(b_k^{\star}\mid x,b_{<k}^{\star}\right)
    -\sum_{k=1}^{N}\log
    p_{\theta}^{\mathrm{Y}}\!\left(y_k^{\star}\mid x,b_{\le k}^{\star}\right).
$}
\end{equation}

\noindent
A standard autoregressive head supervises one next-token target at each
position, whereas a fork in the document tree has multiple continuations. Each
layout node $B_k$
forks into the next layout node $B_{k+1}$ and its content branch $Y_k$; we retain
the former as the main stream and prepend a branch-start token to the latter.
This token opens an independent continuation under the same ancestor context,
and one such token can be added for every extra branch at a higher-arity fork.
Training therefore remains a standard next-token prediction problem while
exposing the parallel graph structure.

%% file: sec/3_3_infer.tex
\input{tab/layout}

\subsection{Inference Speedup}
\label{sec:inference-speedup}

\paragraph{Masked parallel decoding.}
At inference, \model{} follows the conditional dependencies in
Eq.~\eqref{eq:parallel-decode}. Once a region's layout description is complete,
its content stream and the next layout step become available from the same
preceding context. The decoder therefore forks a content stream from the shared
multimodal and layout prefix while the main layout stream continues. Each
content stream retains the context available at its fork and attends only to
this context and its own causal history. It can observe neither later layout
predictions nor sibling contents, and cannot affect the continuing layout
trajectory. In the logical fused form, an ancestor mask enforces these
constraints so that live streams can be processed jointly and retired
independently. Algorithm~\ref{alg:masked-parallel-decoding} in Appendix~
\ref{sec:masked-parallel-decoding-details} formalizes these semantics.

\paragraph{Evaluated serving realization.}
The reported efficiency results use a request-level
vLLM~\citep{kwon2023efficient} realization rather than physically materializing
the fused ancestor mask. The layout stream and every content stream execute as
independent generation requests in a shared serving runtime. Their
request-specific causal contexts preserve the same branch isolation, while
token-identical multimodal and layout prefixes allow automatic prefix caching to
reuse complete key--value blocks that remain resident. Only the unreused suffix
and branch initialization require additional prefill, and continuous batching
overlaps live content requests with the continuing layout request. This
realization requires neither a draft model nor auxiliary decoding heads.
However, its wall-clock cost includes request admission, residual prefill,
scheduling overhead, and imbalance among branches of unequal length. We
therefore base all efficiency claims on measured throughput and end-to-end
latency in the \emph{Serving Efficiency} section. Appendix~
\ref{sec:vllm-serving-realization} details the visibility-equivalent mapping.

%% file: tab/layout.tex
\begin{table*}[!t]
\centering
\scriptsize
\setlength{\tabcolsep}{2.2pt}
\resizebox{\textwidth}{!}{%
\begin{tabular}{llc*{6}{cccc}}
\toprule
\multirow{2}{*}{\textbf{Paradigm}} &
\multirow{2}{*}{\textbf{Method}} &
\multirow{2}{*}{\textbf{Size}} &
\multicolumn{4}{c}{\textbf{Overall}} & \multicolumn{4}{c}{\textbf{Text}}
& \multicolumn{4}{c}{\textbf{Image}} & \multicolumn{4}{c}{\textbf{Table}}
& \multicolumn{4}{c}{\textbf{Formula}} & \multicolumn{4}{c}{\textbf{Full page}} \\
\cmidrule(lr){4-7}\cmidrule(lr){8-11}\cmidrule(lr){12-15}
\cmidrule(lr){16-19}\cmidrule(lr){20-23}\cmidrule(lr){24-27}
& & &
\textbf{IoU} & \textbf{F1} & \textbf{P} & \textbf{R} &
\textbf{IoU} & \textbf{F1} & \textbf{P} & \textbf{R} &
\textbf{IoU} & \textbf{F1} & \textbf{P} & \textbf{R} &
\textbf{IoU} & \textbf{F1} & \textbf{P} & \textbf{R} &
\textbf{IoU} & \textbf{F1} & \textbf{P} & \textbf{R} &
\textbf{IoU} & \textbf{F1} & \textbf{P} & \textbf{R} \\
\midrule
\multirow{3}{*}{Two-stage} & Dolphin-v2 & 3.8B
& 79.5 & 89.1 & 87.6 & 90.8 & \underline{86.6} & \underline{93.0} & \underline{93.9} & \textbf{92.2}
& 68.5 & 83.2 & 78.2 & \textbf{88.7} & 91.6 & 95.5 & \underline{95.9} & 95.2
& 71.3 & 84.8 & 82.5 & 87.3 & \textbf{88.3} & \textbf{93.8} & \underline{94.8} & \textbf{92.9} \\
 & MinerU2.5-Pro & 1.2B
& \underline{81.3} & \underline{90.2} & \underline{88.8} & \textbf{91.7} & 83.4 & 91.1 & 91.1 & \underline{91.2}
& \textbf{75.0} & \textbf{87.0} & \underline{85.8} & \underline{88.2} & \underline{92.9} & \underline{96.4} & 94.6 & \textbf{98.3}
& \underline{73.9} & \underline{86.4} & \underline{83.7} & \textbf{89.1} & 85.3 & 92.1 & 91.8 & \underline{92.5} \\
 & PaddleOCR-VL~1.5 & 1.0B
& \textbf{83.4} & \textbf{91.2} & \textbf{91.5} & \underline{91.0} & \textbf{87.0} & \textbf{93.2} & \textbf{94.2} & \textbf{92.2}
& \underline{74.2} & \underline{86.5} & \textbf{86.2} & 86.7 & \textbf{94.1} & \textbf{97.0} & \textbf{96.3} & \underline{97.7}
& \textbf{78.2} & \textbf{88.3} & \textbf{89.2} & \underline{87.4} & \underline{87.9} & \underline{93.6} & \textbf{95.0} & 92.3 \\
\midrule
\multirow{6}{*}{End-to-end} & Qianfan-OCR & 4.7B
& 79.4 & 88.9 & 90.4 & 87.4 & 83.9 & 91.1 & 92.2 & 90.1
& 65.9 & 81.8 & 83.4 & 80.3 & 90.8 & 95.0 & 95.2 & 94.9
& \textbf{76.9} & \underline{87.4} & \textbf{91.0} & 84.2 & 84.9 & 91.8 & 93.3 & 90.4 \\
 & DeepSeek-OCR-2 & 3.4B
& \textbf{81.8} & \underline{90.3} & \underline{91.3} & \underline{89.4} & \underline{85.4} & \underline{92.1} & 93.6 & \textbf{90.8}
& \underline{74.8} & \underline{86.5} & \underline{89.1} & 84.1 & 92.0 & 95.8 & 96.0 & \underline{95.6}
& \underline{75.0} & 86.8 & 86.5 & \underline{87.1} & \underline{86.7} & 92.9 & 94.8 & \textbf{91.0} \\
 & dots.ocr & 3.0B
& 80.7 & 89.7 & 89.9 & \textbf{89.6} & \textbf{85.9} & \textbf{92.5} & \textbf{96.2} & 89.0
& \textbf{75.1} & \textbf{87.0} & 86.0 & \underline{88.1} & \underline{92.1} & \underline{96.0} & \underline{96.8} & 95.1
& 69.8 & 83.3 & 80.5 & 86.3 & \textbf{88.1} & \textbf{93.7} & \textbf{96.8} & 90.8 \\
 & HunyuanOCR~1.5 & 1.0B
& -- & -- & -- & -- & 51.9 & 71.8 & 74.0 & 69.8
& -- & -- & -- & -- & -- & -- & -- & --
& -- & -- & -- & -- & -- & -- & -- & -- \\
 & MonkeyOCRv2-B-Parsing & 0.7B
& 77.4 & 87.7 & 86.3 & \textbf{89.6} & 82.4 & 90.2 & 92.0 & 88.5
& 62.5 & 79.3 & 71.4 & \textbf{89.2} & 91.8 & 95.8 & 96.0 & \underline{95.6}
& 73.0 & 85.6 & 86.0 & 85.3 & 85.0 & 91.8 & 93.5 & 90.1 \\
 & \model{} (Ours) & 2.1B
& \underline{81.4} & \textbf{91.1} & \textbf{93.3} & 89.2 & 84.3 & \textbf{92.5} & \underline{94.4} & \underline{90.6}
& 74.1 & \textbf{87.0} & \textbf{92.7} & 81.9 & \textbf{92.7} & \textbf{97.0} & \textbf{97.1} & \textbf{96.8}
& 74.3 & \textbf{88.1} & \underline{88.9} & \textbf{87.3} & 85.9 & \underline{93.1} & \underline{95.4} & \underline{90.9} \\
\bottomrule
\end{tabular}%
}
\caption{Layout analysis on OmniDocBench using countable coverage and
acceptable category matching. All metrics are percentages; Overall
macro-averages text, image, table, and formula, with full-page coverage reported
separately. Bold and underline mark the best and second-best values per
paradigm.}
\label{tab:layout}
\end{table*}

%% file: sec/4_experiments.tex
\input{tab/main_quality}

\section{Experiments}

We evaluate \model{} along two complementary dimensions. First, we measure
layout localization and end-to-end document parsing quality on OmniDocBench.
Second, we evaluate serving efficiency under controlled concurrency on a single
GPU. The main text reports the metrics that directly support these quality and
efficiency claims; additional structural and system analyses are deferred to
Appendix~\ref{sec:additional-analysis}.

\subsection{Experimental Setup}

\paragraph{Benchmark.}
Quality evaluation uses all 1{,}651 pages of OmniDocBench
v1.6~\citep{ouyang2024omnidocbench0}, which spans diverse document types,
languages, layouts, and visual conditions. Serving efficiency uses a fixed
384-page subset obtained by systematic uniform sampling and shared by all
systems and concurrency settings. Each
page provides layout regions, reading order, and region-level content
annotations for text, formulas, tables, and figures.

\paragraph{Training data.}
We curate approximately 11 million document-parsing samples from multiple
open-source datasets, public academic and electronic documents, and in-house
collections, covering full pages and region-level table, formula, and text
crops across diverse document types. The corpus is annotated by
MinerU2.5-Pro~\citep{wang2026mineru2},
PaddleOCR-VL-1.5~\citep{cui2026paddleocr}, and
MOCR~\citep{zheng2026multimodal}; cross-model agreement provides both
high-confidence consensus annotations and a signal for difficulty-aware
sampling. To construct the SFT subset, we cluster ViT
embeddings~\citep{dosovitskiy2020image}, balance the data distribution, remove
near-duplicates, and apply CLIP-similarity-guided sampling across clusters to
preserve visual diversity and long-tail cases, yielding a high-quality subset
of approximately 0.5 million examples.

\paragraph{Training recipe.}
We initialize \model{} from Qwen3-VL-2B~\citep{bai2025qwen30vl}, extending its
vocabulary with specific control tokens and expanding the corresponding
input-embedding and language-model-head dimensions. Training then proceeds in
two stages. First, we perform continual pre-training (CPT) for one epoch on 11
million examples to adapt the original Qwen3-VL-2B model to document parsing.
Second, we perform supervised fine-tuning (SFT) on this 0.5-million-example
subset. Both stages use the next-token objective and
ancestor-attention mask, a global batch size of 256, and a learning-rate
schedule comprising 500 warmup steps followed by cosine decay from a peak of
$5\times 10^{-6}$. Training uses bf16 and FSDP across 128 A800 GPUs with a
maximum sequence length of 16{,}384. The same-backbone sequential baseline
used for the controlled efficiency comparison follows an identical training
configuration and serializes all layout-content records into a single causal
sequence.

\paragraph{Comparison systems.}
Our specialized comparison pool is organized by inference paradigm. Two-stage
systems include PaddleOCR-VL~1.5~\citep{cui2026paddleocr},
MinerU2.5-Pro~\citep{wang2026mineru2},
Dolphin-v2~\citep{feng2026dolphin}, and
GLM-OCR~\citep{duan2026glm0ocr}. End-to-end systems include
dots.ocr~\citep{li2025dots0ocr0}, Qianfan-OCR~\citep{dong2026qianfan},
MonkeyOCRv2-B-Parsing~\citep{liu2026monkeyocrv2},
DeepSeek-OCR-2~\citep{wei2026deepseek0ocr}, and
HunyuanOCR~1.5~\citep{li2026hunyuanocr15}. For end-to-end parsing quality, we
additionally report Ovis2.6-30B-A3B~\citep{lu2026ovisocr2},
Gemini 3 Pro,
Gemini 3 Flash,
Qwen3-VL-235B~\citep{bai2025qwen30vl}, GPT-5.2, and
InternVL3.5-241B~\citep{wang2025internvl3050}, using the published OmniDocBench v1.6
Full results from MinerU2.5-Pro~\citep{wang2026mineru2}. The same-backbone
Sequential SFT baseline is used only for the controlled serving-efficiency
comparison; external systems provide quality and efficiency context across
model scales and parsing paradigms.

\begin{figure*}[tbh]
\centering
\includegraphics[width=\textwidth]{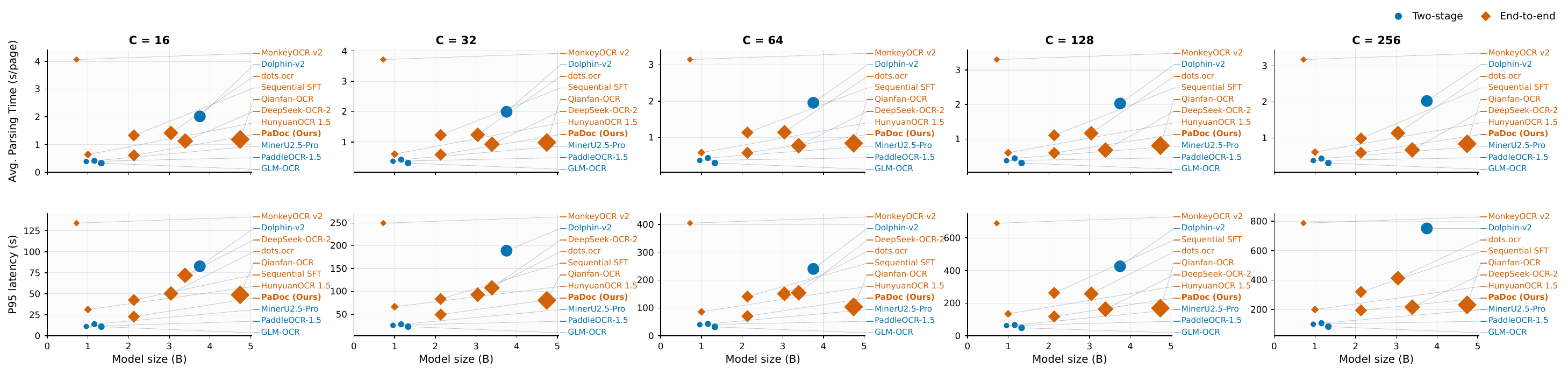}
\caption{Serving efficiency across five concurrency levels on one A800 GPU.
Each point denotes a system positioned by model size, with colors distinguishing
two-stage and end-to-end parsers; the top and bottom rows report mean parsing
time and P95 end-to-end latency, respectively. \model{} attains the lowest mean
parsing time and P95 latency among end-to-end parsers at every concurrency
level, and approaches compact two-stage systems despite using a larger model.}
\label{fig:serving-efficiency}
\end{figure*}

\input{tab/serving_efficiency}

\subsection{Document Parsing Quality}

\paragraph{Evaluation Protocol.}
Following coverage-based layout evaluation~\citep{niu2025mineru2}, we report
classwise IoU, F1, precision, and recall under our countable--acceptable PageIoU
protocol; Appendix~\ref{sec:pageiou-details} specifies its category mapping and
aggregation, and Appendix~\ref{sec:pageiou-additional-results} reports the
remaining protocol variants. For end-to-end parsing, we use the OmniDocBench
v1.6 protocol~\citep{ouyang2024omnidocbench0}, reporting Text Edit, Formula CDM,
Table TEDS/TEDS-S, Read Order Edit, and the standard Overall aggregate of text,
table, and formula quality. All systems parse the same pages without
ground-truth layouts, and cross-scale or cross-paradigm comparisons are
contextual.

\paragraph{Results.}
Table~\ref{tab:layout} shows that \model{} achieves top-tier layout accuracy
among end-to-end systems, with an Overall IoU/F1 of $81.4/91.1$ and the highest
Overall precision ($93.3$). Its advantage is particularly pronounced for
structured regions, where it leads all four table metrics, including $92.7$ IoU
and $97.0$ F1. It also ties the best image F1 ($87.0$) and obtains the best
full-page F1 and recall ($88.1$ and $87.3$) within the end-to-end group. Despite
using a unified model, \model{} remains competitive with specialized two-stage
systems in Overall F1 while attaining the highest Overall precision in the
table, indicating that layout-grounded branching preserves localization
quality. For end-to-end parsing, Table~\ref{tab:main-quality} likewise places
\model{} in the top tier with an Overall score of $94.24$, exceeding Qianfan-OCR
by $0.34$ points and the strongest evaluated general VLM by $0.62$ points. It
attains the best end-to-end Text Edit ($0.038$) and Formula CDM ($95.59$), while
delivering strong Table TEDS/TEDS-S ($90.94/93.39$). These results demonstrate
that parallel decoding preserves competitive quality across layout localization
and regional content recognition.

\subsection{Inference Efficiency}

\paragraph{Evaluation Protocol.}
We serve the autoregressive component of every measured system with vLLM on one
NVIDIA A800 80GB GPU, retaining the official layout detection, cropping, and
post-processing stages of two-stage systems. A closed-loop load generator
maintains 16, 32, 64, 128, or 256 concurrent document requests. Each
model--concurrency setting runs on one isolated GPU replica; settings may run on
different cluster nodes, but no request uses tensor parallelism or cross-node
communication. Table~\ref{tab:serving-efficiency} reports mean valid pages per
second per GPU and P95 end-to-end latency over valid requests. Output-token
throughput, additional latency percentiles, validity statistics, and full
aggregation details are deferred to
Appendix~\ref{sec:parallel-serving-details}.

\paragraph{Implementation Details.}
The \model{} service uses vLLM~0.19.1 with the FlashInfer attention backend,
bfloat16 precision for both model weights and the key--value cache, and one
model replica per GPU without tensor parallelism. The layout stream and each
content branch are submitted as independent requests to the same serving
runtime; this evaluated backend does not materialize the logical ancestor mask.
The runtime supports sequences of up to 32{,}768 tokens and processes at most
384 sequences in each scheduling iteration, subject to a per-iteration budget
of 24{,}576 tokens. It reserves 90\% of the GPU memory for model execution. Both
the layout stream and each content branch have an 8{,}192-token generation limit, and a
document may instantiate at most 255 content branches. Automatic prefix caching
is enabled, whereas speculative decoding and optional cascade-attention
modifications are disabled. We retain per-request latency, scheduler state, and
GPU telemetry traces for all runs.

\paragraph{Results.}
Across every tested concurrency, \model{} has the highest valid-page throughput
and lowest P95 latency among end-to-end parsers. Relative to the same-backbone
Sequential SFT baseline, it improves throughput by $67.4$--$118\%$ and reduces
P95 latency by $39.2$--$54.9\%$. Its throughput reaches 1.722 pages per second
per GPU at
64 concurrent requests and remains near this level at higher concurrency.
Despite using 2.1B parameters, \model{} consistently surpasses the 1.0B
HunyuanOCR~1.5 in both efficiency metrics and substantially outpaces the 0.7B
MonkeyOCRv2-B-Parsing. Figure~\ref{fig:serving-efficiency} further shows that it
approaches compact 1.0--1.3B two-stage systems, indicating that the proposed
parallelism remains effective at a larger model scale. Cross-architecture
comparisons remain contextual, while the same-backbone comparison isolates the
effect of parallel decoding. Appendix~\ref{sec:additional-analysis} provides
additional efficiency analyses.

%% file: tab/main_quality.tex
\begin{table*}[!t]
\centering
\scriptsize
\resizebox{\textwidth}{!}{%
\begin{tabular}{llccccccc}
\toprule
\textbf{Paradigm} & \textbf{Method} & \textbf{Size} &
\textbf{Overall} $\uparrow$ &
$\text{\textbf{Text}}^{\text{\textbf{Edit}}}\downarrow$ &
$\text{\textbf{Formula}}^{\text{\textbf{CDM}}}\uparrow$ &
$\text{\textbf{Table}}^{\text{\textbf{TEDS}}}\uparrow$ &
$\text{\textbf{Table}}^{\text{\textbf{TEDS-S}}}\uparrow$ &
$\text{\textbf{Read Order}}^{\text{\textbf{Edit}}}\downarrow$ \\
\midrule
\multirow{6}{*}{General VLMs}
 & InternVL3.5-241B & 241B & 83.61 & 0.130 & 89.52 & 74.35 & 79.78 & 0.215 \\
 & Qwen3-VL-235B & 235B & 89.78 & \underline{0.063} & 92.53 & 83.07 & 86.75 & 0.166 \\
 & Ovis2.6-30B-A3B & 30B & \textbf{93.62} & \textbf{0.035} & 94.93 & \textbf{89.44} & 92.40 & \textbf{0.135} \\
 & Gemini 3 Pro & -- & \underline{92.85} & 0.064 & \textbf{95.83} & 89.15 & \underline{92.96} & \underline{0.165} \\
 & Gemini 3 Flash & -- & 92.58 & 0.066 & \underline{95.03} & \underline{89.29} & \textbf{93.51} & 0.173 \\
 & GPT-5.2 & -- & 86.52 & 0.114 & 88.00 & 82.95 & 87.93 & 0.193 \\
\midrule
\multirow{4}{*}{Two-stage}
 & Dolphin-v2 & 3.8B & 89.34 & 0.069 & 90.53 & 84.40 & 87.44 & 0.150 \\
 & GLM-OCR & 1.3B & \underline{95.15} & 0.044 & \underline{96.99} & \underline{92.83} & \underline{95.39} & 0.133 \\
 & MinerU2.5-Pro & 1.2B & \textbf{95.69} & \textbf{0.036} & \textbf{97.29} & \textbf{93.42} & \textbf{95.92} & \textbf{0.120} \\
 & PaddleOCR-VL~1.5 & 1.0B & 94.87 & \underline{0.038} & 96.69 & 91.67 & 94.37 & \underline{0.130} \\
\midrule
\multirow{6}{*}{End-to-end}
 & Qianfan-OCR & 4.7B & 93.90 & 0.040 & \underline{95.08} & 90.53 & 93.31 & \underline{0.130} \\
 & DeepSeek-OCR-2 & 3.4B & 90.17 & 0.050 & 91.59 & 83.89 & 87.75 & 0.144 \\
 & dots.ocr & 3.0B & 90.50 & 0.048 & 89.12 & 87.18 & 90.58 & 0.138 \\
 & HunyuanOCR~1.5 & 1.0B & \textbf{94.74} & \underline{0.039} & 94.50 & \textbf{93.67} & \textbf{94.71} & \textbf{0.129} \\
 & MonkeyOCRv2-B-Parsing & 0.7B & 91.57 & 0.053 & 91.83 & 88.24 & 91.38 & 0.131 \\
 & \model{} (Ours) & 2.1B & \underline{94.24} & \textbf{0.038} & \textbf{95.59} & \underline{90.94} & \underline{93.39} & 0.133 \\
\bottomrule
\end{tabular}%
}
\caption{End-to-end document parsing quality on OmniDocBench v1.6 Full.
Where available, published external results are taken from
MinerU2.5-Pro~\citep{wang2026mineru2}, HunyuanOCR-1.5~\citep{li2026hunyuanocr15},
and MonkeyOCRv2~\citep{liu2026monkeyocrv2}. Bold and underline mark the best
and second-best values per paradigm.}
\label{tab:main-quality}
\end{table*}

%% file: tab/serving_efficiency.tex
\begin{table*}[!t]
\centering
\scriptsize
\resizebox{\textwidth}{!}{%
\begin{tabular}{llc|ccccc|ccccc}
\toprule
\multirow{2}{*}{\textbf{Paradigm}} & \multirow{2}{*}{\textbf{Method}} &
\multirow{2}{*}{\textbf{Size}} &
\multicolumn{5}{c|}{\textbf{Mean valid pages per second per GPU} $\uparrow$} &
\multicolumn{5}{c}{\textbf{P95 E2E latency (s)} $\downarrow$} \\
\cmidrule(lr){4-8}\cmidrule(lr){9-13}
& & & \textbf{C16} & \textbf{C32} & \textbf{C64} & \textbf{C128} &
\textbf{C256} & \textbf{C16} & \textbf{C32} & \textbf{C64} &
\textbf{C128} & \textbf{C256} \\
\midrule
\multirow{4}{*}{Two-stage}
& Dolphin-v2 & 3.8B & 0.496 & 0.501 & 0.511 & 0.493 & 0.494 & 82.756 & 189.176 & 240.183 & 426.346 & 751.033 \\
& GLM-OCR & 1.3B & \textbf{3.009} & \textbf{3.248} & \textbf{3.323} & \textbf{3.289} & \textbf{3.280} & \textbf{10.977} & \textbf{22.750} & \textbf{32.341} & \textbf{49.473} & \textbf{83.669} \\
& MinerU2.5-Pro & 1.2B & 2.406 & 2.392 & 2.275 & 2.274 & 2.335 & 13.872 & 27.644 & 43.020 & 66.427 & 106.966 \\
& PaddleOCR-VL~1.5 & 1.0B & \underline{2.560} & \underline{2.727} & \underline{2.701} & \underline{2.692} & \underline{2.691} & \underline{11.219} & \underline{25.537} & \underline{40.151} & \underline{62.861} & \underline{99.979} \\
\midrule
\multirow{7}{*}{End-to-end}
& Qianfan-OCR & 4.7B & 0.844 & 1.017 & 1.184 & 1.238 & 1.193 & 48.856 & 80.313 & 104.076 & 169.998 & 231.610 \\
& DeepSeek-OCR-2 & 3.4B & 0.885 & 1.078 & 1.287 & 1.479 & 1.499 & 71.969 & 107.803 & 154.397 & 163.577 & 216.016 \\
& dots.ocr & 3.0B & 0.705 & 0.808 & 0.872 & 0.857 & 0.881 & 50.387 & 92.664 & 151.081 & 257.744 & 412.459 \\
& HunyuanOCR~1.5 & 1.0B & \underline{1.560} & \underline{1.662} & \underline{1.703} & \underline{1.661} & \underline{1.645} & \underline{31.304} & \underline{66.284} & \underline{86.408} & \underline{135.388} & \underline{199.034} \\
& MonkeyOCRv2-B-Parsing & 0.7B & 0.246 & 0.269 & 0.318 & 0.303 & 0.315 & 133.965 & 249.574 & 404.335 & 689.949 & 787.126 \\
\cmidrule(l){2-13}
& Sequential SFT (Baseline) & 2.1B & 0.751 & 0.814 & 0.879 & 0.904 & 1.015 & 42.610 & 83.176 & 141.042 & 263.247 & 319.337 \\
& \model{} (Ours) & 2.1B
& \makecell{\textbf{1.640}\\[-4.5pt]{\color{PaDocDeepGreen}\fontsize{4.7}{4.7}\selectfont\bfseries (+118\%)}}
& \makecell{\textbf{1.711}\\[-4.5pt]{\color{PaDocDeepGreen}\fontsize{4.7}{4.7}\selectfont\bfseries (+110\%)}}
& \makecell{\textbf{1.722}\\[-4.5pt]{\color{PaDocDeepGreen}\fontsize{4.7}{4.7}\selectfont\bfseries (+95.9\%)}}
& \makecell{\textbf{1.684}\\[-4.5pt]{\color{PaDocDeepGreen}\fontsize{4.7}{4.7}\selectfont\bfseries (+86.3\%)}}
& \makecell{\textbf{1.699}\\[-4.5pt]{\color{PaDocDeepGreen}\fontsize{4.7}{4.7}\selectfont\bfseries (+67.4\%)}}
& \makecell{\textbf{22.847}\\[-4.5pt]{\color{PaDocDeepGreen}\fontsize{4.7}{4.7}\selectfont\bfseries (-46.4\%)}}
& \makecell{\textbf{49.125}\\[-4.5pt]{\color{PaDocDeepGreen}\fontsize{4.7}{4.7}\selectfont\bfseries (-40.9\%)}}
& \makecell{\textbf{70.659}\\[-4.5pt]{\color{PaDocDeepGreen}\fontsize{4.7}{4.7}\selectfont\bfseries (-49.9\%)}}
& \makecell{\textbf{118.808}\\[-4.5pt]{\color{PaDocDeepGreen}\fontsize{4.7}{4.7}\selectfont\bfseries (-54.9\%)}}
& \makecell{\textbf{194.018}\\[-4.5pt]{\color{PaDocDeepGreen}\fontsize{4.7}{4.7}\selectfont\bfseries (-39.2\%)}} \\[-.2em]
\bottomrule
\end{tabular}%
}
\caption{Single-A800 serving efficiency; C16, for example, denotes 16 concurrent
document requests.
Green parentheses show changes from Sequential SFT; Appendix~\ref{sec:parallel-serving-details}
reports additional metrics. Bold/underline: best/second-best per paradigm.}
\label{tab:serving-efficiency}
\end{table*}

%% file: sec/5_conclusion.tex
\section{Conclusion}

We presented \textsc{PaDoc}, a layout-grounded document parser that factorizes
regional content generation under a shared page-image and layout prefix,
exposing region-level parallelism without repeated visual prefills. Packed
variable-length ancestor attention trains this factorization with a standard
next-token objective, while concurrent serving reuses shared prefixes. On one
A800 GPU, \textsc{PaDoc} is the fastest end-to-end parser at every tested
concurrency level, improving throughput by 67.4--118\% and reducing P95 latency
by 39.2--54.9\% relative to the same-backbone Sequential SFT baseline. Despite
its 2.1B size, it also
surpasses the 1.0B HunyuanOCR-1.5, demonstrating that layout-grounded
parallelism can substantially reduce serving cost without modifying the MLLM
architecture.

%% file: sec/A_derivations.tex
\section{Derivations}
\label{sec:derivations}

\subsection{From Region-Specific Sufficiency to Prefix-Conditioned Sufficiency}
\label{sec:proof-prefix-cond}

We provide a derivation of the implication used in the main text:
Assumption~\ref{asm:region-specific} (region-specific content sufficiency),
\begin{equation}
    I\!\left(
        Y_k;\,B_{<k},B_{>k},Y_{<k}
        \mid X,B_k
    \right)
    = 0,
\end{equation}
implies the prefix-conditioned form
\begin{equation}
    I\!\left(
        Y_k;\,B_{>k},Y_{<k}
        \mid X,B_{\le k}
    \right)
    = 0.
\end{equation}
The argument relies on the chain rule of conditional mutual information. For any
random variables $A,B,C$ and conditioning $Z$,
\begin{equation}
    I(A;B,C\mid Z)
    =
    I(A;C\mid Z) + I(A;B\mid Z,C).
\end{equation}
Applying this with $A=Y_k$, $B=(B_{>k},Y_{<k})$, $C=B_{<k}$, $Z=(X,B_k)$ gives
\begin{equation}
\begin{aligned}
    &I(Y_k;B_{<k},B_{>k},Y_{<k}\mid X,B_k)                              \\
    &\qquad=
    I(Y_k;B_{<k}\mid X,B_k)                                             \\
    &\qquad\quad+
    I(Y_k;B_{>k},Y_{<k}\mid X,B_k,B_{<k}).
\end{aligned}
\end{equation}
Since the conditioning $(X,B_k,B_{<k})$ is identical, as a tuple, to
$(X,B_{\le k})$, this becomes
\begin{equation}
\begin{aligned}
    &I(Y_k;B_{<k},B_{>k},Y_{<k}\mid X,B_k)                              \\
    &\qquad=
    I(Y_k;B_{<k}\mid X,B_k)                                             \\
    &\qquad\quad+
    I(Y_k;B_{>k},Y_{<k}\mid X,B_{\le k}).
\end{aligned}
\end{equation}
Conditional mutual information is non-negative, so each term on the right-hand
side is at least zero. Since the left-hand side is zero by assumption, both
terms on the right-hand side must vanish. In particular,
\begin{equation}
    I(Y_k;B_{>k},Y_{<k}\mid X,B_{\le k}) = 0,
\end{equation}
which is the prefix-conditioned form used by \model{}.

The same statement, expressed in conditional-independence notation, is exactly
the weak-union property:
\begin{equation}
\begin{aligned}
    &Y_k \perp\!\!\!\perp (B_{<k},B_{>k},Y_{<k}) \mid X,B_k             \\
    &\qquad\implies
    Y_k \perp\!\!\!\perp (B_{>k},Y_{<k}) \mid X,B_{\le k}.
\end{aligned}
\end{equation}
Note that the converse direction does not hold in general: the prefix-conditioned
form does not, by itself, recover the full region-specific form, because it does
not constrain the dependence of $Y_k$ on $B_{<k}$ given $(X,B_k)$. This is why
\model{} only requires the strictly weaker prefix-conditioned consequence.

%% file: sec/B_pageiou.tex
\section{Details of PageIoU}
\label{sec:pageiou-details}

Inspired by the coverage-based layout evaluation in MinerU2.5
\citep{niu2025mineru2}, we define a PageIoU protocol tailored to fair
cross-model comparison rather than directly adopting their metric. Document
parsers use heterogeneous category inventories and may assign the same region
to categories at different levels of granularity. A hard category
correspondence can therefore conflate taxonomy mismatch with localization
error. We instead map predictions into a common set of broad classes and split
the ground-truth taxonomy into core \emph{include} categories and ambiguous
\emph{acceptable} categories.

The metric compares coverage over the complete page without one-to-one box
matching or an IoU threshold. It also separates binary coverage from countable
coverage, the latter retaining overlapping-box multiplicity. We deliberately do
not use a strict hard-matching mode: such a score would penalize semantically
valid outputs merely because two parsers name or group regions differently.

\subsection{Coverage Representation}

For every page, ground-truth and prediction coordinates are independently
normalized by their respective canvas width and height, giving the common
domain $\Omega=[0,1]^2$. OmniDocBench quadrilaterals are converted to
axis-aligned rectangles using their minimum and maximum coordinates. For broad
class $q$, let $\mathcal{G}^{q}_{\mathrm{inc}}$ be the ground-truth rectangles
whose categories are core members of $q$,
$\mathcal{G}^{q}_{\mathrm{acc}}$ those with acceptable categories, and
$\mathcal{P}^{q}$ the predicted rectangles mapped to $q$. Their pointwise
coverage counts are
\begin{equation}
\begin{aligned}
    a_q(z) &= \sum_{r\in\mathcal{G}^{q}_{\mathrm{inc}}}
              \mathbf{1}[z\in r], \\
    b_q(z) &= \sum_{r\in\mathcal{G}^{q}_{\mathrm{acc}}}
              \mathbf{1}[z\in r], \\
    c_q(z) &= \sum_{r\in\mathcal{P}^{q}}
              \mathbf{1}[z\in r].
\end{aligned}
\end{equation}
Include and acceptable sets are disjoint at the category level, although their
rectangles may overlap spatially.

In \emph{binary} mode, the three counts are clipped pointwise,
\begin{equation}
    (a_q,b_q,c_q)
    \leftarrow
    (\min\{a_q,1\},\min\{b_q,1\},\min\{c_q,1\}),
\end{equation}
so only the union of the boxes in each group matters. In \emph{countable} mode,
the original counts are retained, which penalizes duplicate or overlapping
predictions. We omit $q$ below and write
$\langle f\rangle=\int_{\Omega}f(z)\,\mathrm{d}z$ for area-weighted summation
over the page.

\subsection{Soft Category Modes}

Each category mode defines an intersection $I$, union $U$, and separate
ground-truth and prediction denominators $D_{\mathrm{gt}}$ and
$D_{\mathrm{pr}}$.

\paragraph{Acceptable.}
Acceptable coverage is elastic: it is credited when predicted but does not
create a false negative when omitted. Pointwise, the effective ground-truth
coverage can take any value from $a$ to $a+b$ and is selected to be as
consistent with $c$ as that interval permits:
\begin{equation}
\begin{aligned}
    I_{\mathrm{acc}}
        &= \langle\min\{a+b,c\}\rangle, \\
    U_{\mathrm{acc}}
        &= \langle\max\{a,c\}\rangle, \\
    D^{\mathrm{acc}}_{\mathrm{gt}}
        &= \left\langle\max\{a,\min\{a+b,c\}\}\right\rangle, \\
    D^{\mathrm{acc}}_{\mathrm{pr}}
        &= \langle c\rangle.
\end{aligned}
\end{equation}
Consequently, predicting up to the available acceptable coverage is rewarded in
the intersection, while unpredicted acceptable coverage is absent from the
union and recall denominator.

\paragraph{Ignore acceptable.}
As a robustness protocol, this mode removes locations covered exclusively by
acceptable ground-truth regions from the IoU computation. Define
\begin{equation}
u_{\mathrm{ign}}(a,b,c)=
\begin{cases}
    0, & a=0\ \text{and}\ b>0,\\
    \max\{a,c\}, & \text{otherwise}.
\end{cases}
\end{equation}
The resulting area sums are
\begin{equation}
\begin{aligned}
    I_{\mathrm{ign}}
        &= \langle\min\{a,c\}\rangle, \\
    U_{\mathrm{ign}}
        &= \langle u_{\mathrm{ign}}(a,b,c)\rangle, \\
    D^{\mathrm{ign}}_{\mathrm{gt}}
        &= \langle a\rangle, \\
    D^{\mathrm{ign}}_{\mathrm{pr}}
        &= \left\langle
            c-\min\{\max\{c-a,0\},b\}
           \right\rangle.
\end{aligned}
\end{equation}
The prediction denominator pardons at most $b$ units of prediction coverage
beyond the core coverage $a$. If include and acceptable rectangles overlap
spatially, that location is not exclusively acceptable and remains in the IoU
union. This distinction is visible only in countable mode and preserves its
penalty for overlapping coverage.

\subsection{Per-Page Metrics and Aggregation}

Assuming $0\le x\le y$, as guaranteed by the metric constructions above, we
define a ratio that treats agreement on an empty support as perfect,
\begin{equation}
    \rho(x,y)=
    \begin{cases}
        1, & y=0,\\
        x/y, & y>0.
    \end{cases}
\end{equation}
For every page, class, category mode, and count mode, the metrics are
\begin{equation}
\begin{aligned}
    \mathrm{IoU} &= \rho(I,U), \\
    \mathrm{Precision} &= \rho(I,D_{\mathrm{pr}}), \\
    \mathrm{Recall} &= \rho(I,D_{\mathrm{gt}}).
\end{aligned}
\end{equation}
A page is a support page for class $q$ if it contains at least one include box,
acceptable box, or mapped prediction for that class. Pages with none of the
three are omitted. If $\mathcal{S}_q$ is the support-page set and $m_{pq}$ is a
per-page metric, its reported value is the page-level macro average
\begin{equation}
    \overline{m}_q
    = \frac{1}{|\mathcal{S}_q|}
      \sum_{p\in\mathcal{S}_q} m_{pq}.
\end{equation}
This prevents pages with large canvases or regions from dominating the score.
The reported F1 is derived from macro-averaged precision and recall,
\begin{equation}
    \mathrm{F1}_q
    = \frac{2\overline{P}_q\,\overline{R}_q}
           {\overline{P}_q+\overline{R}_q},
\end{equation}
and is set to zero when both aggregated values are zero.

The primary result in Table~\ref{tab:layout} uses countable coverage with
acceptable category handling. This choice preserves tolerance to category
ambiguity while penalizing overlapping predictions. The remaining three
binary/countable and acceptable/ignore-acceptable combinations are reported in
Section~\ref{sec:pageiou-additional-results}.

\input{tab/pageiou_category_map}

\subsection{Category Mapping}

Table~\ref{tab:pageiou-category-map} specifies the soft mapping from
OmniDocBench ground-truth categories to the five evaluation classes.

\subsection{Exact Computation}

The default backend computes all area sums exactly for axis-aligned rectangles
using coordinate compression. It collects $0$, $1$, and every horizontal and
vertical boundary from the include, acceptable, and prediction boxes. These
coordinates partition $\Omega$ into rectangular cells on which $a$, $b$, and
$c$ are constant. A two-dimensional difference array followed by prefix sums
computes the coverage counts, and each cell contributes its exact area
$\Delta x\Delta y$ to the equations above. Boxes are clipped to $\Omega$ before
the partition is constructed.

We additionally implement a raster backend as a numerical cross-check. It
rounds normalized box boundaries onto a configurable grid, set to
$1000\times1000$ for OmniDocBench, and applies the same equations pixel by
pixel. The raster resolution affects only this checking backend; all reported
results use coordinate-compressed exact areas. Pages are aligned by image
basename, and a missing prediction page is evaluated as an empty layout.

\input{tab/pageiou_binary_acceptable}

\input{tab/pageiou_countable_ignore}

\input{tab/pageiou_binary_ignore}

\subsection{Additional PageIoU Results}
\label{sec:pageiou-additional-results}

Tables~\ref{tab:pageiou-binary-acceptable}--\ref{tab:pageiou-binary-ignore}
report the three protocols not used for the headline layout results. They use
the same per-class IoU, F1, precision, and recall reporting and the same Overall
macro-aggregation as Table~\ref{tab:layout}.

%% file: tab/pageiou_category_map.tex
\begin{table*}[!t]
\centering
\small
\renewcommand{\arraystretch}{0.9}
\begin{tabular*}{0.9\textwidth}{@{\extracolsep{\fill}}lccccc@{}}
\toprule
\textbf{Category} & \textbf{Text} & \textbf{Table} & \textbf{Formula} &
\textbf{Image} & \textbf{Full page} \\
\midrule
Text block                    & $\checkmark$ & $-$          & $-$          & $-$          & $\checkmark$ \\
Title                         & $\checkmark$ & $-$          & $-$          & $-$          & $\checkmark$ \\
Reference                     & $\checkmark$ & $-$          & $-$          & $-$          & $\checkmark$ \\
List group                    & $\checkmark$ & $-$          & $-$          & $-$          & $\checkmark$ \\
Code text                     & $\checkmark$ & $-$          & $-$          & $-$          & $\checkmark$ \\
Code-text caption             & $\checkmark$ & $-$          & $-$          & $-$          & $\checkmark$ \\
Header                        & $\circ$      & $-$          & $-$          & $-$          & $\circ$ \\
Footer                        & $\circ$      & $-$          & $-$          & $-$          & $\circ$ \\
Page number                   & $\circ$      & $-$          & $-$          & $-$          & $\circ$ \\
Page footnote                 & $\circ$      & $-$          & $-$          & $-$          & $\circ$ \\
Table caption                 & $\circ$      & $\circ$      & $-$          & $-$          & $\circ$ \\
Table footnote                & $\circ$      & $\circ$      & $-$          & $-$          & $\circ$ \\
Figure caption                & $\circ$      & $-$          & $-$          & $\circ$      & $\circ$ \\
Figure footnote               & $\circ$      & $-$          & $-$          & $\circ$      & $\circ$ \\
Equation caption              & $\circ$      & $-$          & $\circ$      & $-$          & $\circ$ \\
Algorithm mask                & $\circ$      & $-$          & $-$          & $-$          & $\circ$ \\
Text mask                     & $\circ$      & $-$          & $-$          & $-$          & $\circ$ \\
Table                         & $-$          & $\checkmark$ & $-$          & $-$          & $\checkmark$ \\
Table mask                    & $-$          & $\circ$      & $-$          & $-$          & $\circ$ \\
Isolated equation             & $-$          & $-$          & $\checkmark$ & $-$          & $\checkmark$ \\
Semantic equation             & $-$          & $-$          & $\circ$      & $-$          & $\circ$ \\
Equation explanation          & $-$          & $-$          & $\circ$      & $-$          & $\circ$ \\
Organic chemical formula mask & $-$          & $-$          & $\circ$      & $\circ$      & $\circ$ \\
Figure                        & $-$          & $-$          & $-$          & $\checkmark$ & $\checkmark$ \\
Chart mask                    & $-$          & $-$          & $-$          & $\circ$      & $\circ$ \\
Abandon                       & $-$          & $-$          & $-$          & $-$          & $\circ$ \\
Need mask                     & $-$          & $-$          & $-$          & $-$          & $\circ$ \\
Unknown mask                  & $-$          & $-$          & $-$          & $-$          & $\circ$ \\
\bottomrule
\end{tabular*}
\caption{OmniDocBench ground-truth category mapping used by PageIoU. Category
names are presented with spaces and conventional capitalization for
readability. $\checkmark$ denotes an \emph{include} category, $\circ$ denotes an
\emph{acceptable} category, and a dash denotes no membership. Full page
aggregates content-region coverage rather than denoting a single page-sized
rectangle.}
\label{tab:pageiou-category-map}
\end{table*}

%% file: tab/pageiou_binary_acceptable.tex
\begin{table*}[!t]
\centering
\scriptsize
\setlength{\tabcolsep}{2.2pt}
\resizebox{\textwidth}{!}{%
\begin{tabular}{llc*{6}{cccc}}
\toprule
\multirow{2}{*}{\textbf{Paradigm}} &
\multirow{2}{*}{\textbf{Method}} &
\multirow{2}{*}{\textbf{Size}} &
\multicolumn{4}{c}{\textbf{Overall}} & \multicolumn{4}{c}{\textbf{Text}}
& \multicolumn{4}{c}{\textbf{Image}} & \multicolumn{4}{c}{\textbf{Table}}
& \multicolumn{4}{c}{\textbf{Formula}} & \multicolumn{4}{c}{\textbf{Full page}} \\
\cmidrule(lr){4-7}\cmidrule(lr){8-11}\cmidrule(lr){12-15}
\cmidrule(lr){16-19}\cmidrule(lr){20-23}\cmidrule(lr){24-27}
& & &
\textbf{IoU} & \textbf{F1} & \textbf{P} & \textbf{R} &
\textbf{IoU} & \textbf{F1} & \textbf{P} & \textbf{R} &
\textbf{IoU} & \textbf{F1} & \textbf{P} & \textbf{R} &
\textbf{IoU} & \textbf{F1} & \textbf{P} & \textbf{R} &
\textbf{IoU} & \textbf{F1} & \textbf{P} & \textbf{R} &
\textbf{IoU} & \textbf{F1} & \textbf{P} & \textbf{R} \\
\midrule
\multirow{3}{*}{Two-stage} & Dolphin-v2 & 3.8B
& 79.7 & 89.2 & 87.7 & 91.0 & \underline{87.3} & \underline{93.4} & 93.9 & \underline{92.8}
& 68.6 & 83.2 & 78.3 & \textbf{88.7} & 91.6 & 95.5 & \underline{95.9} & 95.2
& 71.3 & 84.9 & 82.5 & 87.3 & \textbf{88.9} & \textbf{94.2} & 94.9 & \textbf{93.5} \\
 & MinerU2.5-Pro & 1.2B
& \underline{82.3} & \underline{90.8} & \underline{89.9} & \textbf{91.8} & 87.2 & 93.3 & \textbf{95.3} & 91.4
& \textbf{75.1} & \textbf{87.0} & \underline{85.9} & \underline{88.2} & \underline{92.9} & \underline{96.4} & 94.6 & \textbf{98.3}
& \underline{73.9} & \underline{86.4} & \underline{83.8} & \textbf{89.2} & \textbf{88.9} & \textbf{94.2} & \textbf{95.6} & 92.8 \\
 & PaddleOCR-VL~1.5 & 1.0B
& \textbf{83.6} & \textbf{91.4} & \textbf{91.5} & \underline{91.2} & \textbf{87.7} & \textbf{93.6} & \underline{94.3} & \textbf{92.9}
& \underline{74.3} & \underline{86.5} & \textbf{86.3} & 86.7 & \textbf{94.1} & \textbf{97.0} & \textbf{96.3} & \underline{97.7}
& \textbf{78.2} & \textbf{88.3} & \textbf{89.3} & \underline{87.4} & \underline{88.6} & \underline{94.0} & \underline{95.2} & \underline{92.9} \\
\midrule
\multirow{6}{*}{End-to-end} & Qianfan-OCR & 4.7B
& 79.6 & 89.0 & 90.5 & 87.6 & 84.6 & 91.6 & 92.4 & 90.8
& 65.9 & 81.9 & 83.5 & 80.3 & 90.8 & 95.0 & 95.2 & 94.9
& \textbf{76.9} & \underline{87.5} & \textbf{91.0} & 84.2 & 85.9 & 92.4 & 93.8 & 91.0 \\
 & DeepSeek-OCR-2 & 3.4B
& \underline{82.0} & \underline{90.4} & \underline{91.3} & \underline{89.5} & 86.0 & 92.5 & 93.7 & \textbf{91.4}
& \underline{74.9} & 86.6 & \underline{89.2} & 84.1 & 92.0 & 95.8 & 96.0 & \underline{95.6}
& \underline{75.0} & 86.8 & 86.5 & \underline{87.1} & 87.4 & 93.3 & 95.0 & \textbf{91.6} \\
 & dots.ocr & 3.0B
& 80.9 & 89.8 & 89.9 & \textbf{89.8} & \underline{86.5} & \underline{92.9} & \underline{96.4} & 89.6
& \textbf{75.2} & \underline{87.0} & 86.0 & \underline{88.1} & \underline{92.1} & \underline{96.0} & \underline{96.8} & 95.1
& 69.8 & 83.3 & 80.5 & 86.3 & \textbf{88.7} & \underline{94.1} & \underline{97.0} & \underline{91.3} \\
 & HunyuanOCR~1.5 & 1.0B
& -- & -- & -- & -- & 53.0 & 72.7 & 75.3 & 70.2
& -- & -- & -- & -- & -- & -- & -- & --
& -- & -- & -- & -- & -- & -- & -- & -- \\
 & MonkeyOCRv2-B-Parsing & 0.7B
& 77.6 & 87.8 & 86.3 & \textbf{89.8} & 83.0 & 90.6 & 92.1 & 89.1
& 62.6 & 79.3 & 71.4 & \textbf{89.2} & 91.8 & 95.8 & 96.0 & \underline{95.6}
& 73.0 & 85.7 & 86.0 & 85.4 & 85.7 & 92.1 & 93.6 & 90.7 \\
 & \model{} (Ours) & 2.1B
& \textbf{82.1} & \textbf{91.5} & \textbf{94.1} & 89.2 & \textbf{86.7} & \textbf{93.9} & \textbf{97.1} & \underline{90.9}
& 74.5 & \textbf{87.1} & \textbf{93.0} & 81.9 & \textbf{92.8} & \textbf{97.0} & \textbf{97.2} & \textbf{96.8}
& 74.5 & \textbf{88.2} & \underline{89.0} & \textbf{87.3} & \underline{88.2} & \textbf{94.4} & \textbf{97.8} & \underline{91.3} \\
\bottomrule
\end{tabular}%
}
\caption{Layout analysis on OmniDocBench using binary coverage and acceptable
category matching. All metrics are percentages; Overall macro-averages text,
image, table, and formula, with full-page coverage reported separately. Bold
and underline mark the best and second-best values per paradigm.}
\label{tab:pageiou-binary-acceptable}
\end{table*}

%% file: tab/pageiou_countable_ignore.tex
\begin{table*}[!t]
\centering
\scriptsize
\setlength{\tabcolsep}{2.2pt}
\resizebox{\textwidth}{!}{%
\begin{tabular}{llc*{6}{cccc}}
\toprule
\multirow{2}{*}{\textbf{Paradigm}} &
\multirow{2}{*}{\textbf{Method}} &
\multirow{2}{*}{\textbf{Size}} &
\multicolumn{4}{c}{\textbf{Overall}} & \multicolumn{4}{c}{\textbf{Text}}
& \multicolumn{4}{c}{\textbf{Image}} & \multicolumn{4}{c}{\textbf{Table}}
& \multicolumn{4}{c}{\textbf{Formula}} & \multicolumn{4}{c}{\textbf{Full page}} \\
\cmidrule(lr){4-7}\cmidrule(lr){8-11}\cmidrule(lr){12-15}
\cmidrule(lr){16-19}\cmidrule(lr){20-23}\cmidrule(lr){24-27}
& & &
\textbf{IoU} & \textbf{F1} & \textbf{P} & \textbf{R} &
\textbf{IoU} & \textbf{F1} & \textbf{P} & \textbf{R} &
\textbf{IoU} & \textbf{F1} & \textbf{P} & \textbf{R} &
\textbf{IoU} & \textbf{F1} & \textbf{P} & \textbf{R} &
\textbf{IoU} & \textbf{F1} & \textbf{P} & \textbf{R} &
\textbf{IoU} & \textbf{F1} & \textbf{P} & \textbf{R} \\
\midrule
\multirow{3}{*}{Two-stage} & Dolphin-v2 & 3.8B
& 77.6 & 88.0 & 85.7 & 90.6 & \underline{80.9} & \underline{89.8} & \underline{88.0} & \underline{91.6}
& 68.0 & 82.8 & 77.7 & \textbf{88.7} & 91.0 & 95.2 & \underline{95.3} & 95.2
& 70.5 & 84.4 & 81.9 & 87.0 & \textbf{87.7} & \textbf{93.5} & \underline{94.4} & \textbf{92.7} \\
 & MinerU2.5-Pro & 1.2B
& \underline{79.5} & \underline{89.2} & \underline{87.1} & \textbf{91.4} & 77.8 & 87.9 & 85.6 & 90.5
& \textbf{74.2} & \textbf{86.5} & \underline{85.1} & \underline{87.9} & \underline{92.6} & \underline{96.2} & 94.3 & \textbf{98.3}
& \underline{73.2} & \underline{86.0} & \underline{83.2} & \textbf{88.9} & 84.9 & 91.9 & 91.5 & \underline{92.2} \\
 & PaddleOCR-VL~1.5 & 1.0B
& \textbf{81.5} & \textbf{90.2} & \textbf{89.7} & \underline{90.8} & \textbf{81.2} & \textbf{90.1} & \textbf{88.6} & \textbf{91.7}
& \underline{73.6} & \underline{86.2} & \textbf{85.6} & 86.7 & \textbf{93.2} & \textbf{96.6} & \textbf{95.5} & \underline{97.7}
& \textbf{77.8} & \textbf{88.1} & \textbf{89.1} & \underline{87.1} & \underline{87.5} & \underline{93.4} & \textbf{94.8} & 92.0 \\
\midrule
\multirow{6}{*}{End-to-end} & Qianfan-OCR & 4.7B
& 77.3 & 87.8 & 88.5 & 87.2 & 78.0 & 87.9 & 86.3 & 89.6
& 65.2 & 81.6 & 82.8 & 80.3 & 89.6 & 94.4 & 94.0 & 94.9
& \textbf{76.4} & \underline{87.2} & \textbf{90.8} & 83.9 & 84.4 & 91.5 & 92.9 & 90.1 \\
 & DeepSeek-OCR-2 & 3.4B
& \textbf{80.0} & \underline{89.4} & \underline{89.6} & 89.2 & \textbf{80.8} & \textbf{89.8} & \underline{89.2} & \textbf{90.4}
& \underline{74.2} & 86.2 & \underline{88.5} & 84.1 & 90.9 & 95.3 & 95.0 & \underline{95.6}
& \underline{74.3} & 86.4 & 86.0 & \underline{86.8} & \underline{86.4} & 92.7 & 94.6 & \textbf{90.9} \\
 & dots.ocr & 3.0B
& 78.8 & 88.7 & 88.1 & \underline{89.3} & \underline{80.1} & \underline{89.5} & \textbf{90.7} & 88.3
& \textbf{74.8} & \textbf{86.8} & 85.6 & \underline{88.1} & \underline{91.4} & \underline{95.7} & \textbf{96.2} & 95.1
& 69.0 & 82.8 & 79.9 & 85.9 & \textbf{87.7} & \textbf{93.4} & \textbf{96.6} & 90.5 \\
 & HunyuanOCR~1.5 & 1.0B
& -- & -- & -- & -- & 49.4 & 70.0 & 71.8 & 68.3
& -- & -- & -- & -- & -- & -- & -- & --
& -- & -- & -- & -- & -- & -- & -- & -- \\
 & MonkeyOCRv2-B-Parsing & 0.7B
& 75.5 & 86.7 & 84.5 & \textbf{89.4} & 76.8 & 87.1 & 86.5 & 87.8
& 62.2 & 79.1 & 71.0 & \textbf{89.2} & 90.9 & 95.3 & 95.1 & \underline{95.6}
& 72.2 & 85.1 & 85.4 & 84.9 & 84.5 & 91.4 & 93.1 & 89.7 \\
 & \model{} (Ours) & 2.1B
& \underline{79.3} & \textbf{89.9} & \textbf{91.1} & 88.9 & 78.5 & 89.2 & 88.6 & \underline{89.9}
& 73.6 & \underline{86.7} & \textbf{92.1} & 81.9 & \textbf{91.5} & \textbf{96.0} & \underline{95.3} & \textbf{96.8}
& 73.7 & \textbf{87.7} & \underline{88.4} & \textbf{86.9} & 85.5 & \underline{92.8} & \underline{95.0} & \underline{90.6} \\
\bottomrule
\end{tabular}%
}
\caption{Layout analysis on OmniDocBench using countable coverage and
ignore-acceptable category matching. All metrics are percentages; Overall
macro-averages text, image, table, and formula, with full-page coverage reported
separately. Bold and underline mark the best and second-best values per
paradigm.}
\label{tab:pageiou-countable-ignore}
\end{table*}

%% file: tab/pageiou_binary_ignore.tex
\begin{table*}[!t]
\centering
\scriptsize
\setlength{\tabcolsep}{2.2pt}
\resizebox{\textwidth}{!}{%
\begin{tabular}{llc*{6}{cccc}}
\toprule
\multirow{2}{*}{\textbf{Paradigm}} &
\multirow{2}{*}{\textbf{Method}} &
\multirow{2}{*}{\textbf{Size}} &
\multicolumn{4}{c}{\textbf{Overall}} & \multicolumn{4}{c}{\textbf{Text}}
& \multicolumn{4}{c}{\textbf{Image}} & \multicolumn{4}{c}{\textbf{Table}}
& \multicolumn{4}{c}{\textbf{Formula}} & \multicolumn{4}{c}{\textbf{Full page}} \\
\cmidrule(lr){4-7}\cmidrule(lr){8-11}\cmidrule(lr){12-15}
\cmidrule(lr){16-19}\cmidrule(lr){20-23}\cmidrule(lr){24-27}
& & &
\textbf{IoU} & \textbf{F1} & \textbf{P} & \textbf{R} &
\textbf{IoU} & \textbf{F1} & \textbf{P} & \textbf{R} &
\textbf{IoU} & \textbf{F1} & \textbf{P} & \textbf{R} &
\textbf{IoU} & \textbf{F1} & \textbf{P} & \textbf{R} &
\textbf{IoU} & \textbf{F1} & \textbf{P} & \textbf{R} &
\textbf{IoU} & \textbf{F1} & \textbf{P} & \textbf{R} \\
\midrule
\multirow{3}{*}{Two-stage} & Dolphin-v2 & 3.8B
& 77.8 & 88.1 & 85.7 & 90.8 & 81.6 & 90.1 & 88.0 & \underline{92.3}
& 68.0 & 82.8 & 77.7 & \textbf{88.7} & 91.0 & 95.2 & \underline{95.3} & 95.2
& 70.6 & 84.4 & 81.9 & \underline{87.1} & \underline{88.3} & \textbf{93.9} & 94.5 & \textbf{93.3} \\
 & MinerU2.5-Pro & 1.2B
& \underline{80.5} & \underline{89.8} & \underline{88.1} & \textbf{91.5} & \underline{81.7} & \underline{90.3} & \textbf{89.8} & 90.7
& \textbf{74.3} & \textbf{86.5} & \underline{85.2} & \underline{87.9} & \underline{92.6} & \underline{96.2} & 94.3 & \textbf{98.3}
& \underline{73.3} & \underline{86.0} & \underline{83.2} & \textbf{89.0} & \textbf{88.5} & \textbf{93.9} & \textbf{95.4} & 92.5 \\
 & PaddleOCR-VL~1.5 & 1.0B
& \textbf{81.7} & \textbf{90.3} & \textbf{89.8} & \underline{91.0} & \textbf{81.9} & \textbf{90.5} & \underline{88.7} & \textbf{92.4}
& \underline{73.7} & \underline{86.2} & \textbf{85.7} & 86.7 & \textbf{93.2} & \textbf{96.6} & \textbf{95.5} & \underline{97.7}
& \textbf{77.8} & \textbf{88.1} & \textbf{89.1} & \underline{87.1} & 88.2 & \underline{93.8} & \underline{95.0} & \underline{92.7} \\
\midrule
\multirow{6}{*}{End-to-end} & Qianfan-OCR & 4.7B
& 77.5 & 87.9 & 88.5 & 87.3 & 78.8 & 88.3 & 86.5 & \underline{90.2}
& 65.3 & 81.6 & 82.8 & 80.3 & 89.6 & 94.4 & 94.0 & 94.9
& \textbf{76.5} & \underline{87.2} & \textbf{90.8} & 83.9 & 85.3 & 92.1 & 93.4 & 90.8 \\
 & DeepSeek-OCR-2 & 3.4B
& \textbf{80.3} & \underline{89.5} & \underline{89.7} & \underline{89.4} & \textbf{81.5} & \underline{90.2} & 89.3 & \textbf{91.0}
& \underline{74.3} & 86.3 & \underline{88.6} & 84.1 & 90.9 & 95.3 & 95.0 & \underline{95.6}
& \underline{74.4} & 86.4 & 86.0 & \underline{86.8} & 87.1 & 93.1 & 94.8 & \textbf{91.5} \\
 & dots.ocr & 3.0B
& 79.0 & 88.8 & 88.1 & \textbf{89.5} & 80.7 & 89.8 & \underline{90.8} & 88.9
& \textbf{74.8} & \underline{86.8} & 85.6 & \underline{88.1} & \underline{91.4} & \underline{95.7} & \textbf{96.2} & 95.1
& 69.1 & 82.8 & 79.9 & 85.9 & \textbf{88.3} & \underline{93.8} & \underline{96.8} & \underline{91.0} \\
 & HunyuanOCR~1.5 & 1.0B
& -- & -- & -- & -- & 50.4 & 70.8 & 73.1 & 68.7
& -- & -- & -- & -- & -- & -- & -- & --
& -- & -- & -- & -- & -- & -- & -- & -- \\
 & MonkeyOCRv2-B-Parsing & 0.7B
& 75.7 & 86.8 & 84.5 & \textbf{89.5} & 77.3 & 87.4 & 86.5 & 88.4
& 62.3 & 79.1 & 71.1 & \textbf{89.2} & 90.9 & 95.3 & 95.1 & \underline{95.6}
& 72.2 & 85.2 & 85.4 & 84.9 & 85.1 & 91.8 & 93.2 & 90.3 \\
 & \model{} (Ours) & 2.1B
& \underline{80.1} & \textbf{90.3} & \textbf{91.9} & 89.0 & \underline{80.9} & \textbf{90.7} & \textbf{91.2} & \underline{90.2}
& 73.9 & \textbf{86.9} & \textbf{92.5} & 81.9 & \textbf{91.6} & \textbf{96.1} & \underline{95.4} & \textbf{96.8}
& 73.9 & \textbf{87.8} & \underline{88.6} & \textbf{87.0} & \underline{87.8} & \textbf{94.1} & \textbf{97.4} & \underline{91.0} \\
\bottomrule
\end{tabular}%
}
\caption{Layout analysis on OmniDocBench using binary coverage and
ignore-acceptable category matching. All metrics are percentages; Overall
macro-averages text, image, table, and formula, with full-page coverage reported
separately. Bold and underline mark the best and second-best values per
paradigm.}
\label{tab:pageiou-binary-ignore}
\end{table*}

%% file: sec/C_ablations_of_training_paradigm.tex
\section{Ablations of the Training Paradigm}
\label{sec:training-paradigm-ablation}

The ancestor-attention objective in the Training Paradigm section
admits several systems realizations. They expose the same token visibility and
optimize the same next-token objective, but differ in how they represent and
execute the sparse attention pattern. This appendix defines the three backends
we compared and reports their memory, single-node efficiency, and multi-node
scaling. The comparison is an implementation ablation rather than a change to
the probabilistic factorization or training targets.

\subsection{Attention Backends}

Let a packed sample be $[M\mid Y_1\mid\cdots\mid Y_N]$, where $M$ contains the
shared image and layout stream, and let $M_{\le f_k}$ be the prefix visible at
the fork of branch $Y_k$. The desired attention consists of one causal problem
for $M$ and one causal problem per branch,
\begin{equation}
    Q_M \rightarrow K_M,
    \qquad
    Q_{Y_k} \rightarrow [K_{M_{\le f_k}};K_{Y_k}].
\end{equation}
The three backends implement this same relation as follows.

\paragraph{Dense SDPA.}
Our original implementation materializes a Boolean mask of shape
$B\times S\times S$ during collation and dispatches it through PyTorch scaled
dot-product attention (SDPA). This representation is simple and serves as the
reference implementation, but its arbitrary two-dimensional mask prevents the
text layers from using the standard causal FlashAttention-2 path. Its mask
storage and construction also grow quadratically with the padded sequence
length.

\paragraph{Flex Attention.}
The block-sparse baseline expresses the same ancestor relation as a compiled
block mask for PyTorch Flex Attention. It can skip fully masked query--key
blocks without constructing the dense mask, but introduces graph-compilation
and block-scheduling overhead. The first optimizer step is particularly costly
because it includes compilation; we exclude that cold step from all steady-state
timing comparisons.

\paragraph{Tree-varlen (ours).}
We decompose the packed tree into the variable-length causal problems above and
execute them with the FlashAttention-2 varlen kernel. Queries and Q/K/V
projections are computed once in the original packed order. For each branch, we
gather only the already-projected K/V rows corresponding to its visible shared
prefix and private causal history. The backward pass of the gather sums
gradients from all branches into the original shared-prefix rows. Branch outputs
are scattered back to the packed order. FlashAttention's bottom-right causal
alignment makes a branch query of length $|Y_k|$ see the full prefix and only
non-future tokens in $Y_k$. The gather indices and temporary K/V tensors scale
with the concatenated varlen K/V rows rather than a materialized $S\times S$
mask.

\begin{figure*}[t]
    \centering
    \includegraphics[width=0.98\textwidth]{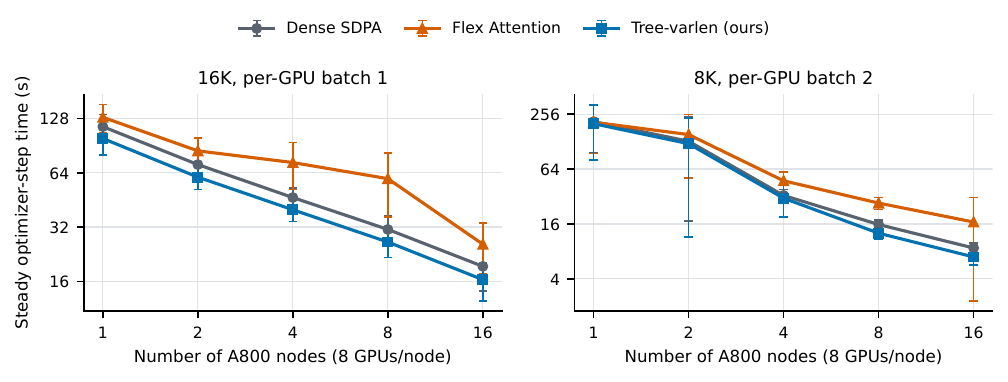}
    \caption{Multi-node training efficiency on A800 GPUs (lower is better).
    Each point is the mean optimizer-step time over steps 2--4, and error bars
    show the sample standard deviation. Global batch size is fixed at 256.
    Dense SDPA and Flex Attention are the two baselines; Flex remains slower
    despite avoiding the dense mask, while tree-varlen is consistently fastest.
    The large error bars for 8K at
    one and two nodes arise from input samples whose prompts already exceed the
    length cap and trigger data rolling.}
    \label{fig:training-backend-scaling}
\end{figure*}

\subsection{Experimental Protocol}

We benchmark the Qwen3-VL-2B backbone used in the main experiments. To isolate
the decoder attention backend, all configurations use a frozen vision encoder,
bf16, FSDP full
sharding, and gradient checkpointing. Each node contains eight A800 GPUs. We
use each configuration's actual end-to-end training path: the dense and Flex
configurations dispatch their named backends through the corresponding model
configuration; in our configuration, tree-varlen is applied to the text decoder
while the frozen vision encoder retains standard FlashAttention-2. We
hold the global batch size at 256 and vary the cluster size over
$\{1,2,4,8,16\}$ nodes. We use two token-budget-matched settings: maximum
sequence length 16{,}384 with per-GPU batch size 1, and maximum sequence length
8{,}192 with per-GPU batch size 2. Accordingly, gradient accumulation is
$\{32,16,8,4,2\}$ for 16K and $\{16,8,4,2,1\}$ for 8K.

Each run executes four optimizer steps on the same training corpus. We discard
step 1 from timing because it contains cold data/kernel startup and, for Flex,
graph compilation. Table~\ref{tab:training-backend-ablation} reports the mean
and sample standard deviation over steps 2--4. A step includes exposed data
loading/collation time, all accumulation microsteps, forward and backward
passes, FSDP communication, gradient clipping, and the optimizer update. We
synchronize CUDA and take the maximum wall time and peak memory over all ranks.
Peak allocated memory covers all four steps, including the cold step.

\input{tab/training_backend_ablation}

\subsection{Results and Backend Selection}

Flex Attention does not provide a favorable trade-off in this setting. It is
slower than SDPA at every 16K point and at 2--16 nodes for 8K. On 16 nodes it
achieves only $0.754\times$ and $0.519\times$ the SDPA training speed for 16K
and 8K, respectively. Its excluded cold step also takes 258.951 seconds at 16K
and 423.558 seconds at 8K because of compilation.

Tree-varlen gives the most consistent efficiency improvement. At 16K it is
$1.163$--$1.185\times$ faster than dense SDPA across all cluster sizes. At 8K,
the gain grows from $1.053\times$ on one node to $1.254\times$ on 16 nodes. At
16 nodes this reduces the optimizer-step time from 19.391 to 16.366 seconds for
16K, and from 8.720 to 6.953 seconds for 8K. Its peak-memory reduction is more
modest: approximately 0.75 GiB for 16K and 0.37 GiB for 8K, indicating that
model state, multimodal activations, and non-attention tensors dominate the
observed peak.

At 8K, the one- and two-node timings
have very high variance because the data pipeline rolls past samples whose
prompt alone exceeds 8K. This creates apparently super-linear transitions and
should not be interpreted as pure communication efficiency. The cleaner
8-to-16-node transition reaches 90.5\% efficiency for SDPA and 80.7\% for Flex,
while tree-varlen reaches 91.2\%. For 16K, tree-varlen sustains 75.5--81.8\%
efficiency at each node doubling, similar to or slightly better than the
baselines. Based on its consistent steady-state speedup, low compilation
overhead, and exact realization of the ancestor mask, we use tree-varlen for
all main experiments.

%% file: tab/training_backend_ablation.tex
\begin{table*}[!t]
\centering
\scriptsize
\resizebox{\textwidth}{!}{%
\begin{tabular}{lc|cc|ccc|ccc}
\hline
& & \multicolumn{2}{c|}{\textbf{Dense SDPA}} &
\multicolumn{3}{c|}{\textbf{Flex Attention}} &
\multicolumn{3}{c}{\textbf{Tree-varlen (ours)}} \\
\textbf{Setting} & \textbf{Nodes} & \textbf{Step (s)} $\downarrow$ &
\textbf{Peak GiB} $\downarrow$ & \textbf{Step (s)} $\downarrow$ &
\textbf{Peak GiB} $\downarrow$ & \textbf{vs. SDPA} $\uparrow$ &
\textbf{Step (s)} $\downarrow$ & \textbf{Peak GiB} $\downarrow$ &
\textbf{vs. SDPA} $\uparrow$ \\
\hline
16K / b1 & 1  & $115.242\!\pm\!19.302$ & 52.341 & $129.851\!\pm\!22.938$ & 51.592 & $0.887\times$ & $\mathbf{99.052\!\pm\!18.953}$ & 51.592 & $\mathbf{1.163\times}$ \\
16K / b1 & 2  & $71.033\!\pm\!9.471$   & 51.419 & $84.611\!\pm\!15.234$  & 50.669 & $0.840\times$ & $\mathbf{60.518\!\pm\!8.828}$   & 50.669 & $\mathbf{1.174\times}$ \\
16K / b1 & 4  & $46.660\!\pm\!6.032$   & 50.764 & $72.899\!\pm\!21.113$  & 50.014 & $0.640\times$ & $\mathbf{39.918\!\pm\!5.537}$   & 50.014 & $\mathbf{1.169\times}$ \\
16K / b1 & 8  & $31.063\!\pm\!6.018$   & 50.436 & $59.353\!\pm\!23.067$  & 49.687 & $0.523\times$ & $\mathbf{26.449\!\pm\!4.699}$   & 49.686 & $\mathbf{1.174\times}$ \\
16K / b1 & 16 & $19.391\!\pm\!5.182$   & 50.272 & $25.711\!\pm\!8.061$   & 49.523 & $0.754\times$ & $\mathbf{16.366\!\pm\!3.943}$   & 49.523 & $\mathbf{1.185\times}$ \\
\hline
8K / b2  & 1  & $212.364\!\pm\!114.703$ & 51.051 & $209.159\!\pm\!114.207$ & 50.682 & $1.015\times$ & $\mathbf{201.591\!\pm\!120.474}$ & 50.683 & $\mathbf{1.053\times}$ \\
8K / b2  & 2  & $129.408\!\pm\!112.275$ & 49.554 & $153.161\!\pm\!101.817$ & 49.185 & $0.845\times$ & $\mathbf{121.665\!\pm\!110.101}$ & 49.187 & $\mathbf{1.064\times}$ \\
8K / b2  & 4  & $32.877\!\pm\!5.377$    & 48.996 & $47.655\!\pm\!12.306$   & 48.627 & $0.690\times$ & $\mathbf{30.695\!\pm\!11.598}$   & 48.628 & $\mathbf{1.071\times}$ \\
8K / b2  & 8  & $15.786\!\pm\!2.010$    & 48.717 & $27.130\!\pm\!3.985$    & 48.348 & $0.582\times$ & $\mathbf{12.677\!\pm\!1.841}$    & 48.349 & $\mathbf{1.245\times}$ \\
8K / b2  & 16 & $8.720\!\pm\!1.200$     & 45.202 & $16.809\!\pm\!14.539$   & 44.834 & $0.519\times$ & $\mathbf{6.953\!\pm\!1.269}$     & 44.835 & $\mathbf{1.254\times}$ \\
\hline
\end{tabular}%
}
\caption{Training-backend ablation. ``vs. SDPA'' is the dense-SDPA step time
divided by the backend step time at the same sequence length and node count.
Peak memory is maximum allocated CUDA memory over all ranks and all four steps.
The best steady step time in each row is bold.}
\label{tab:training-backend-ablation}
\end{table*}

%% file: sec/D_additional_analysis.tex
\section{Controlled Quality and Structural Efficiency Analyses}
\label{sec:additional-analysis}

This appendix complements the cross-model results in the main text with three
controlled analyses. We first isolate the effect of the parallel training
objective under an identical backbone and data configuration. We then quantify
the reduction in autoregressive forward steps, report paired wall-clock
speedups, and examine how the measured speedup changes with page complexity.
Unless otherwise noted, efficiency analyses use the same systematically
sampled 384-page OmniDocBench subset, A800 hardware, vLLM version, and
concurrency levels as the main experiment.

\subsection{Same-Backbone Quality Comparison}

\input{tab/same_backbone_layout}
\input{tab/same_backbone_e2e}

Table~\ref{tab:same-backbone-layout} isolates layout grounding while holding
the Qwen3-VL-2B backbone, training corpus, and PageIoU protocol fixed.
\model{} improves F1 for every reported class: Overall F1 increases by 1.460
points, with the largest gain on formulas (2.231 points). Precision also
increases for every class, including gains of 5.574 points on images and 3.729
points Overall. These improvements are accompanied by lower recall for Overall,
Image, Table, and Full page, indicating a more selective localization profile
rather than uniformly denser predictions.

Table~\ref{tab:same-backbone-e2e} evaluates the resulting regional contents
under the same controlled setting. Aggregate end-to-end quality is preserved,
with an Overall change of only $-0.345$ points; Text Edit and both table metrics
improve, while Reading Order Edit and Formula CDM remain close to the
Sequential SFT baseline. Together, the two tables show that the efficiency gain
does not require sacrificing either layout grounding or regional content
recognition.

\subsection{Forward-Step Reduction and Measured Speedup}

For a page with layout regions $B_{1:N}$ and regional contents
$Y_{1:N}$, the numbers of autoregressive forward steps along a sequential
generation path and the \model{} critical path are
\begin{equation}
\begin{aligned}
    F_{\mathrm{seq}}
        &= \ell(B_{1:N})+\sum_{k=1}^{N}\ell(Y_k),\\
    F_{\model}
        &= \max_{1\le k\le N}
           \left[\ell(B_{\le k})+\ell(Y_k)\right].
\end{aligned}
\end{equation}
We summarize the per-page reduction as
$R_{\mathrm{step}}=1-F_{\model}/F_{\mathrm{seq}}$, using the layout and branch
traces generated by \model{}. This quantity describes the number of decoding
iterations removed from the autoregressive path; it is not a wall-clock
speedup estimate.

Separately, we evaluate \model{} and our Sequential SFT model on the same pages,
hardware, vLLM version, and document concurrencies. We pair valid requests by
page and compute the measured end-to-end speedup as
$S_{\mathrm{e2e},i}=L^{\mathrm{seq}}_i/L^{\model}_i$. Both systems retain the
serving limits specified in the main experiment.

\input{tab/forward_steps_speedup}

Table~\ref{tab:forward-steps-speedup}(a) shows that, at C64, the median number
of forward steps decreases from 725.0 for a sequentialized response to 297.5
along the \model{} critical path. The median per-page reduction is 58.4\%,
with an interquartile range of 47.7--68.2\%.
Table~\ref{tab:forward-steps-speedup}(b) independently reports wall-clock
behavior: median paired speedup ranges from $1.453\times$ to $1.937\times$
across the five concurrency levels, corresponding to median latency reductions
of 31.2--48.4\%.

%% file: tab/same_backbone_layout.tex
\begin{table*}[!t]
\centering
\scriptsize
\setlength{\tabcolsep}{2.2pt}
\resizebox{\textwidth}{!}{%
\begin{tabular}{l*{6}{cccc}}
\toprule
\multirow{2}{*}{\textbf{Method}} &
\multicolumn{4}{c}{\textbf{Overall}} & \multicolumn{4}{c}{\textbf{Text}}
& \multicolumn{4}{c}{\textbf{Image}} & \multicolumn{4}{c}{\textbf{Table}}
& \multicolumn{4}{c}{\textbf{Formula}} & \multicolumn{4}{c}{\textbf{Full page}} \\
\cmidrule(lr){2-5}\cmidrule(lr){6-9}\cmidrule(lr){10-13}
\cmidrule(lr){14-17}\cmidrule(lr){18-21}\cmidrule(lr){22-25}
&
\textbf{IoU} & \textbf{F1} & \textbf{P} & \textbf{R} &
\textbf{IoU} & \textbf{F1} & \textbf{P} & \textbf{R} &
\textbf{IoU} & \textbf{F1} & \textbf{P} & \textbf{R} &
\textbf{IoU} & \textbf{F1} & \textbf{P} & \textbf{R} &
\textbf{IoU} & \textbf{F1} & \textbf{P} & \textbf{R} &
\textbf{IoU} & \textbf{F1} & \textbf{P} & \textbf{R} \\
\midrule
Sequential SFT
& 80.905 & 89.666 & 89.556 & \textbf{89.860}
& 83.346 & 90.744 & 91.455 & 90.057
& \textbf{74.177} & 86.327 & 87.088 & \textbf{85.745}
& 92.243 & 95.746 & 94.695 & \textbf{96.835}
& 73.854 & 85.848 & 84.984 & 86.804
& 85.692 & 92.322 & 93.078 & \textbf{91.604} \\
\model{} (Ours)
& \textbf{81.355} & \textbf{91.126} & \textbf{93.284} & 89.157
& \textbf{84.258} & \textbf{92.497} & \textbf{94.445} & \textbf{90.628}
& 74.109 & \textbf{86.951} & \textbf{92.661} & 81.904
& \textbf{92.729} & \textbf{96.978} & \textbf{97.127} & 96.830
& \textbf{74.324} & \textbf{88.079} & \textbf{88.905} & \textbf{87.267}
& \textbf{85.941} & \textbf{93.125} & \textbf{95.446} & 90.913 \\
\bottomrule
\end{tabular}%
}
\caption{Controlled layout comparison using the same Qwen3-VL-2B backbone,
training corpus, and countable--acceptable PageIoU protocol. All metrics are
percentages; Overall macro-averages text, image, table, and formula, with
full-page coverage reported separately. Bold marks the better value in each
pair.}
\label{tab:same-backbone-layout}
\end{table*}

%% file: tab/same_backbone_e2e.tex
\begin{table*}[!t]
\centering
\scriptsize
\begin{tabular}{lcccccc}
\toprule
\textbf{Method} &
\textbf{Overall} $\uparrow$ &
$\text{\textbf{Text}}^{\text{\textbf{Edit}}}\downarrow$ &
$\text{\textbf{Read Order}}^{\text{\textbf{Edit}}}\downarrow$ &
$\text{\textbf{Table}}^{\text{\textbf{TEDS}}}\uparrow$ &
$\text{\textbf{Table}}^{\text{\textbf{TEDS-S}}}\uparrow$ &
$\text{\textbf{Formula}}^{\text{\textbf{CDM}}}\uparrow$ \\
\midrule
Sequential SFT
& \textbf{94.585} & 0.03827 & \textbf{0.13014}
& 90.664 & 92.827 & \textbf{96.919} \\
\model{} (Ours)
& 94.240 & \textbf{0.03810} & 0.13320
& \textbf{90.940} & \textbf{93.390} & 95.590 \\
\bottomrule
\end{tabular}
\caption{Controlled end-to-end parsing comparison on OmniDocBench v1.6 Full
using the same Qwen3-VL-2B backbone, training corpus, and evaluation protocol.
Bold marks the better value in each pair.}
\label{tab:same-backbone-e2e}
\end{table*}

%% file: tab/forward_steps_speedup.tex
\begin{table*}[!t]
\centering
\scriptsize
\begin{tabular}{lcccc}
\toprule
\multicolumn{5}{l}{\textbf{(a) Autoregressive forward steps on the C64 paired set}} \\
\textbf{Metric} & \textbf{Mean} &
\textbf{P25} & \textbf{Median} & \textbf{P75} \\
\midrule
$F_{\mathrm{seq}}$ (sequentialized response) & 923.2 & 376.5 & 725.0 & 1270.5 \\
$F_{\model}$ (critical path) & 325.3 & 182.5 & 297.5 & 441.0 \\
Forward-step reduction (\%) & 56.3 & 47.7 & 58.4 & 68.2 \\
\midrule
\multicolumn{5}{l}{\textbf{(b) Measured paired end-to-end speedup}} \\
\textbf{Concurrency} & \textbf{Paired pages} &
\textbf{Sequential SFT P50 (s)} &
\textbf{\model{} P50 (s)} &
\textbf{Speedup P50 (P25--P75)} \\
\midrule
16  & 374 & 13.546  & 8.547  & 1.616 (1.143--2.259) \\
32  & 375 & 27.483  & 15.759 & 1.633 (1.211--2.305) \\
64  & 376 & 49.160  & 32.410 & 1.453 (1.029--2.193) \\
128 & 376 & 96.596  & 65.991 & 1.483 (1.032--2.176) \\
256 & 375 & 198.999 & 99.555 & 1.937 (1.444--2.858) \\
\bottomrule
\end{tabular}
\caption{Forward-step reduction and measured end-to-end speedup on the uniform
384-page subset. Panel (a) summarizes generated traces from 376 valid C64
pages. Panel (b) reports paired \model{}/Sequential SFT latency measurements.
Speedup quantiles are computed from per-page latency ratios and therefore need
not equal ratios of the corresponding marginal latency quantiles.}
\label{tab:forward-steps-speedup}
\end{table*}

%% file: sec/E_additional_serving_metrics.tex
\section{Masked Parallel Decoding and Additional Serving Details}
\label{sec:parallel-serving-details}
\label{sec:additional-serving-metrics}

\subsection{Masked Parallel Decoding}
\label{sec:masked-parallel-decoding-details}

Masked parallel decoding specifies the runtime-independent visibility and fork
semantics of \model{}. For concreteness, consider a logical fused realization
whose key--value cache contains $L$ positions. For $C$ live streams, an ancestor
mask $M\in\{0,1\}^{C\times L}$ records whether stream $c$ may attend to cached
position $\ell$:
\begin{equation}
    M_{c,\ell}=1
    \quad\Longleftrightarrow\quad
    \ell\in\Anc(c)\ \text{or the private history of }c.
\end{equation}
This formulation defines the permitted dependencies; it does not require every
serving backend to materialize $M$ physically.

When one input token from every live stream is appended, the logical state is
updated as
\begin{equation}
    M\leftarrow[\,M\ I_C\,],
    \qquad L\leftarrow L+C.
\end{equation}
The identity block exposes each new token only to its own stream. With additive
bias $A_M$, where masked entries are $-\infty$, a decoder layer computes
\begin{equation}
    \operatorname{Attn}(Q)
    =\operatorname{softmax}\!\left(
      \frac{QK^{\top}}{\sqrt d}+A_M
    \right)V.
\end{equation}
Thus the current tokens of all live streams can be processed in one logical
batched step without exposing their private suffixes to one another.

When the layout stream completes region $B_k$, its boundary output has not yet
been appended to the cache. The scheduler therefore copies the layout stream's
pre-boundary visibility,
\begin{equation}
    M_{\mathrm{child},1:L}
    \leftarrow M_{\mathrm{layout},1:L},
\end{equation}
and creates the corresponding content stream with a learned branch-initialization
input. On the next step, the identity extension gives the layout stream
visibility to its boundary output and the content stream visibility to its
initialization input, but neither stream can observe the other's new input. The
content stream therefore sees $(X,B_{\le k})$ and its own continuation, but
neither future layout nodes nor sibling contents. Algorithm~
\ref{alg:masked-parallel-decoding} summarizes this logical fused realization.

\begin{algorithm}[H]
\caption{Masked parallel decoding in \model{}}
\label{alg:masked-parallel-decoding}
\begin{algorithmic}[1]
\STATE Prefill $X$; initialize the layout stream, cache length $L$, and mask $M$.
\WHILE{any stream is live}
    \STATE Let $C$ be the live-stream count and $\mathbf{i}$ their next inputs.
    \STATE Append $\mathbf{i}$ to the cache: $M\leftarrow[\,M\ I_C\,]$, $L\leftarrow L+C$.
    \STATE Decode $\mathbf{o}$ for all streams using the additive bias $A_M$.
    \STATE Set each existing stream's next input to its output token.
    \IF{the layout output completes region $B_k$}
        \STATE Copy the layout row and create $Y_k$ with a branch-initialization input.
    \ENDIF
    \STATE Retire streams whose output is $\EOS$.
\ENDWHILE
\end{algorithmic}
\end{algorithm}

A completed stream's mask row is removed. Its private cache positions remain
invisible to all surviving streams and need not be revisited. Algorithm~
\ref{alg:masked-parallel-decoding} is a semantic and logical fused formulation;
the reported experiments use the visibility-equivalent request-level mapping
described next.

\subsection{Request-Level vLLM Realization}
\label{sec:vllm-serving-realization}

The serving backend used in our experiments realizes the same visibility with
one request per stream rather than a fused forward pass over $M$. Let $P$ denote
the original multimodal prompt and let $B_{\le k}^{-}$ denote the main-stream
tokens through $B_k$ but before its closing trigger. When the trigger is
observed, the child request concatenates the original multimodal prompt, this
pre-boundary layout prefix, and a learned branch-initialization input. Because
the child request contains only this shared context and its private
continuation, standard causal attention enforces the same branch isolation as
the corresponding row of $M$.

The layout request and all child requests are submitted to the same
shared vLLM serving runtime. Parent and child requests preserve identical
tokenized prefixes, multimodal inputs, and preprocessing settings, and share a
document-specific cache namespace. With automatic prefix caching, matching
complete paged key--value cache blocks that remain resident can be reused by a
child request. Only the suffix after the last reusable block boundary, together
with the branch-initialization input, requires residual child prefill. The child
becomes schedulable only after the step that produced the region boundary;
continuous batching then overlaps it with the continuing layout stream and
other live branches. When a branch finishes, the serving runtime returns its
private cache blocks to the memory pool.

This implementation is visibility-equivalent to Algorithm~
\ref{alg:masked-parallel-decoding}, but its physical execution differs: it does
not materialize the logical ancestor mask or process every stream in one fused
forward pass. Its latency includes request admission, residual prefix
processing, kernel and host overhead, and idle work from unequal branch lengths.
The critical-path ratio derived in the main text is therefore an ideal
structural quantity rather than a wall-clock guarantee; all efficiency claims
use the measured throughput and end-to-end latency reported in the
\emph{Serving Efficiency} section.

\subsection{Additional Serving Metrics}

\paragraph{Protocol and aggregation.}
All results in this subsection use the same fixed 384-page subset as
Table~\ref{tab:serving-efficiency}, shared by every system and concurrency
setting. Each model--concurrency pair is measured in one complete closed-loop
run on a single A800 GPU after two warm-up requests, with no request retry. The
timed path includes the official layout detection, cropping, and post-processing
stages for two-stage systems. A page contributes to valid-page throughput only
when inference completes normally, structured-output validation succeeds, and
neither a generation limit nor an early-stop safeguard is reached. Let
$N_{\mathrm{valid}}$ be the number of such pages, $O_i$ the generated-token
count of request $i$, $T$ the wall time of the complete run, and $G=1$ the GPU
count. The sustained rates are
\begin{equation}
  R_{\mathrm{page}}=\frac{N_{\mathrm{valid}}}{TG},
  \qquad
  R_{\mathrm{tok}}=\frac{\sum_i O_i}{TG}.
\end{equation}
The wall-time denominator includes all work in the run, including work spent on
outputs that are subsequently marked invalid or truncated, whereas
output-token throughput includes every generated token. Following the main
paper, E2E latency statistics are calculated over valid document requests.
We sample vLLM's monotonic generation-token counter every 0.2 seconds. Peak
output-token throughput is the maximum counter increase per second over any
fixed 10-second sliding window.

\paragraph{Validity.}
Across all five concurrency settings, all 384 \model{} requests completed
normally, with no service errors, timeouts, generation-limit truncations, or
early-stop events. The Sequential SFT runs likewise contained no service errors
or timeouts; 359--364 pages per setting satisfied the valid-page criterion,
while the remaining 20--25 outputs reached the generation limit. As defined
above, tokens generated before truncation remain included in output-token
throughput.

\paragraph{Latency distribution.}
Table~\ref{tab:additional-serving-latency} supplements the P95 values
in Table~\ref{tab:serving-efficiency} with the mean, median, P90, and P99 of
valid-request E2E latency for \model{}. The P95 row reproduces the corresponding
\model{} entries in Table~\ref{tab:serving-efficiency}.

\input{tab/additional_serving_latency}

\input{tab/additional_serving_throughput}

\paragraph{Serving envelope.}
All autoregressive calls use a 32{,}768-token context limit and an 8{,}192-token
generation limit. The standard scheduler envelope allows 256 sequences and
32{,}768 batched tokens per iteration. The evaluated \model{} configuration
instead allows 384 sequences with a 24{,}576-token budget, as reported in the
main text. The vLLM memory reservation is 90\%, except for the PaddleOCR-VL
services, which use 80\% to accommodate their official layout stage.

\paragraph{Output-token throughput.}
Table~\ref{tab:additional-serving-throughput} reports mean sustained and peak
output-token throughput for the controlled same-backbone comparison between
Sequential SFT and \model{}. Relative to Sequential SFT, \model{} improves
sustained output throughput by 26.9--44.4\% across all tested concurrency
settings and also achieves higher peak output throughput throughout.

For \model{}, sustained output throughput rises from 1{,}410.0 tokens/s/GPU at
C16 to 1{,}551.2 at C64 and remains within 2.2\% of this value at C128--C256.
This trend agrees with the valid-page throughput in
Table~\ref{tab:serving-efficiency}, which peaks at C64 and varies by at most
2.2\% thereafter. Peak 10-second output throughput increases from 2{,}655.7
tokens/s/GPU at C16 to 3{,}576.5 at C256. Thus C64 is the highest sustained
throughput operating point in the tested range, while higher concurrency
increases short-window token throughput on this workload.

%% file: tab/additional_serving_latency.tex
\begin{table}[H]
\centering
\scriptsize
\resizebox{\columnwidth}{!}{%
\begin{tabular}{lccccc}
\toprule
\textbf{E2E latency (s)} & \textbf{C16} & \textbf{C32} & \textbf{C64} &
\textbf{C128} & \textbf{C256} \\
\midrule
Mean & 10.126 & 18.341 & 35.648 & 67.556 & 105.367 \\
P50  & 8.527  & 15.674 & 31.938 & 65.235 & 99.555 \\
P90  & 17.938 & 32.541 & 66.546 & 102.118 & 161.978 \\
P95  & 22.847 & 49.125 & 70.659 & 118.808 & 194.018 \\
P99  & 39.583 & 60.223 & 82.079 & 142.508 & 210.073 \\
\bottomrule
\end{tabular}
}
\caption{Valid-request E2E latency (s) for \model{} on the fixed 384-page
serving workload.}
\label{tab:additional-serving-latency}
\end{table}

%% file: tab/additional_serving_throughput.tex
\begin{table}[H]
\centering
\scriptsize
\resizebox{\columnwidth}{!}{%
\begin{tabular}{llccccc}
\toprule
\textbf{Statistic} & \textbf{Method} & \textbf{C16} & \textbf{C32} &
\textbf{C64} & \textbf{C128} & \textbf{C256} \\
\midrule
\multirow{2}{*}{\textbf{Mean}}
& Sequential SFT (Baseline) & 997.1 & 1{,}066.3 & 1{,}222.2 & 1{,}171.9 & 1{,}141.3 \\
& \model{} (Ours) & \textbf{1{,}410.0} & \textbf{1{,}539.2} & \textbf{1{,}551.2} & \textbf{1{,}517.9} & \textbf{1{,}528.7} \\
\midrule
\multirow{2}{*}{\textbf{Peak}}
& Sequential SFT (Baseline) & 1{,}539.5 & 1{,}986.6 & 2{,}253.1 & 2{,}899.3 & 3{,}542.4 \\
& \model{} (Ours) & \textbf{2{,}655.7} & \textbf{2{,}883.5} & \textbf{2{,}895.8} & \textbf{3{,}410.8} & \textbf{3{,}576.5} \\
\bottomrule
\end{tabular}%
}
\caption{Mean and peak output-token throughput (\mbox{TPS/GPU}) on the uniform
384-page workload. C16, for example, denotes 16 concurrent document requests;
peak rates use a fixed 10-second window.}
\label{tab:additional-serving-throughput}
\end{table}